\documentclass{article}

    \PassOptionsToPackage{numbers, compress}{natbib}
\usepackage[preprint]{neurips_2024}

\usepackage{natbib}
\usepackage[utf8]{inputenc} % allow utf-8 input
\usepackage[T1]{fontenc}    % use 8-bit T1 fonts
\usepackage{hyperref}       % hyperlinks
\usepackage{url}            % simple URL typesetting
\usepackage{booktabs}       % professional-quality tables
\usepackage{amsfonts}       % blackboard math symbols
\usepackage{nicefrac}       % compact symbols for 1/2, etc.
\usepackage{microtype}      % microtypography
\usepackage{xcolor}         % colors

\usepackage{adjustbox}
\usepackage{ulem}
\usepackage{multirow} 
\usepackage{amsmath,amsfonts,bm}

\def\eqref#1{equation~\ref{#1}}
\def\1{\bm{1}}

\DeclareMathAlphabet{\mathsfit}{\encodingdefault}{\sfdefault}{m}{sl}
\SetMathAlphabet{\mathsfit}{bold}{\encodingdefault}{\sfdefault}{bx}{n}

\DeclareMathOperator*{\argmax}{arg\,max}

\usepackage[capitalize,noabbrev]{cleveref}
\usepackage{amsmath}
\usepackage{amssymb}
\usepackage{mathtools}
\usepackage{amsthm}

\newcommand{\ie}{\textit{i}.\textit{e}.}

\usepackage{color}
\usepackage{subfigure}
\usepackage{graphicx}
\usepackage{enumitem}

\usepackage{tikz}
\usepackage{pgfplots}
\usepackage{colortbl}
\pgfplotsset{compat=1.16}
\usepackage{amssymb}
\usepackage{wrapfig}
\usepackage{hyperref}
\usepackage{algorithm}
\usepackage{algorithmic}

\title{Contrastive Diffuser: Planning Towards High Return States via Contrastive Learning}

\author{%
  Yixiang Shan  \\
  Jilin University\\
  China\\
  \And
  Zhengbang Zhu  \\
  Shanghai Jiaotong University\\
  China\\
  \And
  Ting Long\thanks{Corresponding author.}  \\
  Jilin University\\
  China\\
  \texttt{longting@jlu.edu.cn} \\
  \And
  Qifan Liang \\
  Jilin University\\
  China\\
  \And
  Yi Chang\thanks{Corresponding author.}  \\
  Jilin University\\
  China\\
  \texttt{yichang@jlu.edu.cn} \\
  \And
  Weinan Zhang \\
  Shanghai Jiaotong University\\
  China\\
  \And
  Liang Yin \\
  Shanghai Jiaotong University\\
  China\\
}

\begin{document}

\maketitle

\begin{abstract}

The performance of offline reinforcement learning (RL) is sensitive to the proportion of high-return trajectories in the offline dataset. However, in many simulation environments and real-world scenarios, there are large ratios of low-return trajectories rather than high-return trajectories, which makes learning an efficient policy challenging. In this paper, we propose a method called Contrastive Diffuser (CDiffuser) to make full use of low-return trajectories and improve the performance of offline RL algorithms. Specifically, CDiffuser groups the states of trajectories in the offline dataset into high-return states and low-return states and treats them as positive and negative samples correspondingly. Then, it designs a contrastive mechanism to pull the trajectory of an agent toward high-return states and push them away from low-return states. Through the contrast mechanism, trajectories with low returns can serve as negative examples for policy learning, guiding the agent to avoid areas associated with low returns and achieve better performance. Experiments on 14 commonly used D4RL benchmarks demonstrate the effectiveness of our proposed method. Our code is publicly available at \url{https://anonymous.4open.science/r/CDiffuser}.

\end{abstract}

% \vspace{-20pt}
\section{Introduction}
\label{sec:intro}
Offline reinforcement learning (offline RL)~\citep{levine2020offline,prudencio2023survey} is a significant branch of reinforcement learning, where an agent is trained on pre-collected offline datasets and is evaluated online. Since offline RL avoids potential risks from interacting with the environment during policy learning, it has broad applications in numerous real-world scenarios, like commercial recommendation~\citep{xiao2021general}, health care~\citep{fatemi2022semi}, dialog systems~\citep{jaques2020human}, and autonomous driving~\citep{shi2021offline}.

\begin{figure*}
    
    \vspace{-19pt}
    \centering
    \includegraphics[width=1\textwidth]{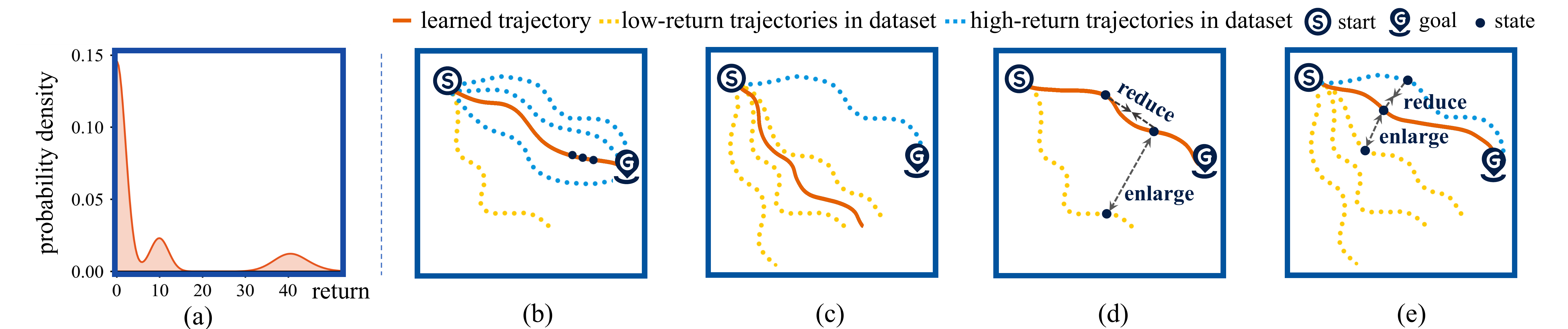}
    \vspace{-10pt}
    \caption{(a) The probability density of trajectories’ returns in Maze2d; (b) The learned trajectory when high-return trajectories are abundant;  (c) The learned trajectory when the number of high-return trajectories is limited; (d) The contrastive learning applied by previous RL models; (e) The example of our solution.
    }
    \vspace{-19pt}
    \label{fig:intro}
\end{figure*}

% \lt{
However, the performance of offline RL methods highly depends on the proportion of the high-return trajectories in the offline dataset.
% , since they are trained with the offline data. 
When the dataset contains a large proportion of high-return trajectories, as is presented in Figure \ref{fig:intro}(b), offline RL methods can easily learn the pattern of high-return trajectories such that they can achieve excellent performance when interacting with the environment.
In contrast, when the dataset has a limited number of high-return trajectories, as is presented in Figure \ref{fig:intro}(c), offline RL methods struggle to learn a good pattern from the dataset to achieve high returns.
Unfortunately, the issue of limited high-return trajectories commonly exists in both simulation environments (\textit{e.g.}, Maze2d) and real-world scenarios (\textit{e.g.}, robotics control and medical diagnosis). As it is illustrated in Figure \ref{fig:intro}(a), we visualize the probability density of trajectories' returns in Maze2d. We can observe that the number of high-return trajectories is much limited.

% \lt{
Although the number of high-return trajectories is limited in aforementioned cases, 
there are abundant low-return trajectories. 
Notably, few works consider developing techniques to make full use of low-return trajectories.
As states in those low-return trajectories indicate potential areas where agents might obtain low returns,
% . Generally speaking, if 
we assume that if an agent achieves relatively high returns during its interactions with the environment, the trajectories generated during the interaction should maintain some different states from the low-return trajectories in the offline dataset. 
As illustrated in Figure \ref{fig:intro}(a), some states (marked in blue dot) in the trajectories that can reach the goal position
% in the trajectory of success to the goal position 
have a clear boundary from the ones in low-return trajectories.
Therefore, as it is illustrated in Figure \ref{fig:intro}(e), 
one conservative solution to enable offline RL algorithms to achieve high returns is taking advantage of the state differences between high-return and low-return trajectories: keeping the states in the agent's trajectory close to high-return states, and away from low-return states. 

% To achieve this goal, we need to address two issues: (1) what techniques to adopt; (2) which attributes of the trajectory to apply the techniques to. For the first issue, 
However, there are no mature techniques to pull the states of a trajectory toward high-return states and push them away from low-return states, to the best of our knowledge. Fortunately, 
% in the light of contrastive learning,
there is an analogous case: contrastive learning,
which aims to bring 
a given sample close to positive ($i.e.$, similar) samples and far from negative ($i.e.$, dissimilar) samples~\citep{xiao2020should, tian2020makes, wang2022contrastive, khosla2020supervised}. Inspired by that, we propose to treat states with high return in trajectories of offline dataset as positive samples and those with low return as negative samples, and leverage contrastive learning to pull the states toward high-return states and push them away from low-return states, as Figure\ref{fig:intro}(d) illustrates. It is worth noting that, unlike previous works~\citep{qiu2022contrastive,laskin2020curl,yuan2022robust,agarwal2020contrastive}, which apply contrastive learning to constrain the states of the same trajectory to similar representations and the states of different trajectories to dissimilar representations, as is illustrated in Figure \ref{fig:intro}(c), we aim to \textbf{use contrastive learning to constrain policy toward high-return states and away from low-return states}.  
Furthermore, the criteria for distinguishing positive and negative samples here are based on the returns rather than the labels.

Through the contrast mechanism, trajectories with low returns can serve as negative examples for policy learning, guiding the agent to avoid areas associated with low returns. Additionally, with the guidance of high-return states, the agent ultimately achieves high returns.
However, ordinary states are feedback from the environment rather than generated by the model, applying contrastive mechanisms to these states produces no gradient for policy optimization.
Considering some diffusion-based RL methods generate subsequent trajectories for planning \citep{janner2022planning,ajay2023is}, in which abundant states are generated by policy model, we build our constrastive mechanism on those diffusion-based RL methods and propose a method called \textbf{Contrastive Diffuser} (\textbf{CDiffuser}). Specifically, we first group the states of the trajectories in the offline dataset into high-return states and low-return states. 
Then, we learn a diffusion-based trajectory generation model to generate the subsequent trajectories, and apply a contrastive mechanism to constrain the states of the generated trajectories by pulling them toward the high-return states and pushing them away from the low-return states in the offline dataset.
With the contrastive mechanism constrained states for planning, the agent makes decisions towards the high-return states.
To evaluate the performance of CDiffuser, we conduct experiments on 14 D4RL~\citep{fu2020d4rl} benchmarks. The experiment results demonstrate that CDiffuser has superior performance.

In summary, our contributions are: 
(\romannumeral 1) We propose a method called CDiffuser, which takes the advantage of low-return trajectories by pulling the states in trajectories toward to high-return states and pushing them away from low-return states.
(\romannumeral 2) We perform contrastive learning to constrain the states in the agent's trajectory and enhance the policy learning.
To the best of our knowledge, our work is the first which apply contrastive learning to enhance the policy learning.
(\romannumeral 3) Experiment results on 14 D4RL datasets demonstrate the outstanding performance of CDiffuser.

\vspace{-10pt}
\section{Preliminaries}
\vspace{-6pt}

\subsection{Denoising Probabilistic Models}
\vspace{-6pt}
Denoising Probabilistic Models (Diffusion Models)~\citep{sohl2015deep,songdenoising,ho2020denoising} are a group of generative models, which generate samples by denoising from Gaussian noise.
A diffusion model is composed of a forward process and a backward process.
Given the original data $\bm{x} \sim q(\bm{x})$, the forward process transfers $\bm{x}$ into Gaussian noise by gradually adding noise, \ie, $
    q(\bm{x}^i \vert \bm{x}^{i-1}) = \mathcal{N}(\bm{x}^i; \sqrt{1 - {\beta}^i} \bm{x}^{i-1}, {\beta}^i\bm{I})$, in which $\bm{I}$ is an identity matrix, ${\beta}^i$ is the noise schedule measuring the proportion of noise added at each step, $\bm{x}^0:=\bm{x}$ is a sample from the offline dataset, $\bm{x}^1, \bm{x}^2, ...$ are the latents of diffusion. The backward process recovers $\bm{x}$ by gradually removing the noise at each step, which is formulated with a Gaussian distribution~\citep{Feller1949OnTT} parameterized by $\theta$, \ie, $
p_{\theta}(\bm{x}^{i-1}|\bm{x}^i)=\mathcal{N}({\mu}_{\theta}(\bm{x}^i, i), {\Sigma}_{\theta}(\bm{x}^i, i))$, 
where ${\mu}_\theta(\bm{x}^i, i )=\frac{\sqrt{\alpha^i}(1-\bar{\alpha}^{i})}{1-\bar{\alpha}^{i-1}}\bm{x}^i + \frac{\sqrt{\bar{\alpha}^{i-1}}\beta^i}{1-\bar{\alpha}^i}\psi_\theta(\bm{x}^i, i)$,
$\bar{\alpha}^i = \prod_{j=1}^{i} (1-\beta^i)$ and $\psi_{\theta}(\cdot,\cdot)$ is a model to reconstruct $\bm{x}$.
The objective function can be formulated as follows if we fix ${\Sigma}_{\theta}(\bm{x}^i, t) = \beta_i\bm{I}$~\citep{ho2020denoising}:
\begin{equation}
    \mathcal{L} = \mathbb{E}_{\bm{x}^0,\, i \sim [1, N]}\left[\| \bm{x}^0 - \psi_\theta(\bm{x}^i,i) \|^2 \right].
\end{equation}

\vspace{-6pt}
\subsection{Contrastive Learning}
\vspace{-6pt}
%Contrastive representation learning methods [17, 46, 53, 54,61, 77, 84, 86, 87, 122, 124, 129] take as input pairs of positive and negative examples, and learn representations so that positive pairs have similar representations and negative pairs have dissimilar representations.
Contrastive learning ~\citep{schroff2015facenet,sohn2016improved,khosla2020supervised, yeh2022decoupled,oord2018representation} is a class of self-supervised learning methods which aim at pulling similar samples together and pushing dissimilar samples away from each other.
% embedding augmented versions of the same sample close to each other while trying to push away embeddings from different samples. 
% Therefore\shanyx{Specifically}, given a sample \textbf{x}, its similar samples $\textbf{S}^{+}$ denotes as positive samples, and its different samples denotes as 
% the positive samples $\textbf{S}^{x+}$ which has similarity with \textbf{x}, and the negative samples which dissimilar with $\textbf{x}$, the learning target of contrastive learning is minimizing the distances between $\textbf{x}$ and $\textbf{S}^{+}$, and maximizing the distances between $\textbf{x}$ and $\textbf{S}^{-}$, \ie,
Specifically, given a sample $\bm{x}$ and a similarity measure, 
%the positive sample ${\bm{x}}^{+}$ is defined as the sample similar to $\bm{x}$, and
the positive set $\mathcal{S}^{+}$ is defined as the collection of samples similar to $\bm{x}$, while 
the negative set $\mathcal{S}^{-}$ is defined as the collection of samples dissimilar to $\bm{x}$. Contrastive learning minimizes the distance of between $\bm{x}$ and $\mathcal{S}^{+}$, and maximizes the distance of $\bm{x}$ and $\mathcal{S}^{-}$. That is, for each sample $\bm{x}$, select a positive sample ${\bm{x}}^{+} \in \mathcal{S}^{+}$ and negative samples ${\bm{x}}^{-} \in \mathcal{S}^{-}$. As such, the learning loss is:
\begin{equation}
        \mathcal{L}=-\log\left[\frac{\text{exp}( \text{sim}(f(\bm{x}), f(\bm{x}^{+})) )}{ \text{exp}( \text{sim}(f(\bm{x}), f(\bm{x}^{+})) ) + \sum_{\bm{x}^{-}\in\mathcal{S}^{-}} \text{exp}( \text{sim}(f(\bm{x}), f(\bm{x}^{-})) )  }\right],
\label{eq:infonce}
\end{equation}
where $f(\cdot)$ is the function to map samples to a latent space and sim($\cdot, \cdot$) is the similarity measure.

\subsection{Offline RL Problem Definition}
% \zhengbang{We should highlight offline RL, i.e., learning from a fixed dataset.}

% \blue{The general framework of Reinforcement Learning (RL) mainly includes three parts: policy, actor and environment. At each moment $k$, policy provides action $a_k$ based on the current environment state $s_k$, the actor takes action $a_k$ over environment, and then the environment will give rewards $r_k$ based on the action taken and trans to a new state $s_{k+1}$.
% agent与环境交互的过程组成trajectory
% given THE offline dataset D composed by a groups of traj, the learning target of offline reinforcement learning is to learn a policy which maximizes the reward when deploy the agetn interact with env, \ie
% }

% \longting{
% The general framework of Reinforcement Learning (RL) mainly includes 
\label{sec:preliminary:rl}
Considering a system composed of three parts: policy, agent, and environment. 
The environment in RL is usually formulated as a Markov Decision Process (MDP)~\citep{sutton2018reinforcement} $\mathcal{M} = \{ \mathcal{S}, \mathcal{A}, \mathcal{P}, r, \gamma \}$, where $\mathcal{S}$ is the state space, $\mathcal{A}$ is the action space, $\mathcal{P}(s' | s,a)$ is the transition function, $\gamma$ represents the discount factor, $r$ is the instant reward of each step.
At each step $t$, the agent responds to the state of environment $\bm{s}_t$ by action $\bm{a}_t$ according to policy  
% $\pi$, 
$\pi_\theta$ parameterized by $\theta$, 
and gets an instant return $r_t$. The interaction history is formulated as a trajectory 
% $\tau = \{(\textbf{s}_t, \textbf{a}_t) | t \geq 0\}$. 
$\bm{\tau} = \{(\bm{s}_t, \bm{a}_t, r_t) | t \geq 0\}$. 
% \longting{$\tau_\theta = \{(\bm{s}_t, \bm{a}_t) | t \geq 0\}^{\mathcal{D}}$. }
In this paper, we define the cumulative discounted reward from step $t$ as $v_t = \sum_{i\geq t}\eta^{i-t}r_i$ and call it the return of $\bm{s}_t$.  
%\shanyx{Note that the discount factor $\eta$ is carefully selected to confirm $\eta^{t_{max}}$ stays close to 0, where $t_{max}$ represents the maximum episode length of environment.}

We focus on the offline RL setting in this paper. Therefore, given an offline dataset
% collected by a behavior policy 
$\mathcal{D}\triangleq\{(\bm{s}_t, \bm{a}_t, {r}_t, \bm{s}_{t+1})| t \geq 0\}$ consisting of transition tuples, and 
defining the return of trajectory $\bm{\tau}$ as $R(\bm{\tau}) \triangleq \sum_{t\geq0}\gamma^tr_t$, our goal is learning 
$\pi_\theta$ to maximize the expected return without directly interacting with the environment, \ie, 
% a policy $\pi_\theta$ parameterized by $\theta$ which maximises the expected return, \ie, 
\begin{equation}
    \pi_\theta = \argmax_\theta\mathbb{E}_{\bm{\tau}\sim\pi_\theta} [R(\bm{\tau})] ~.
\end{equation}

\vspace{-10pt}
\section{Methodology}
\vspace{-5pt}

\begin{figure*}
    \centering
    \vspace{-25pt}
    \includegraphics[width=0.9\textwidth]{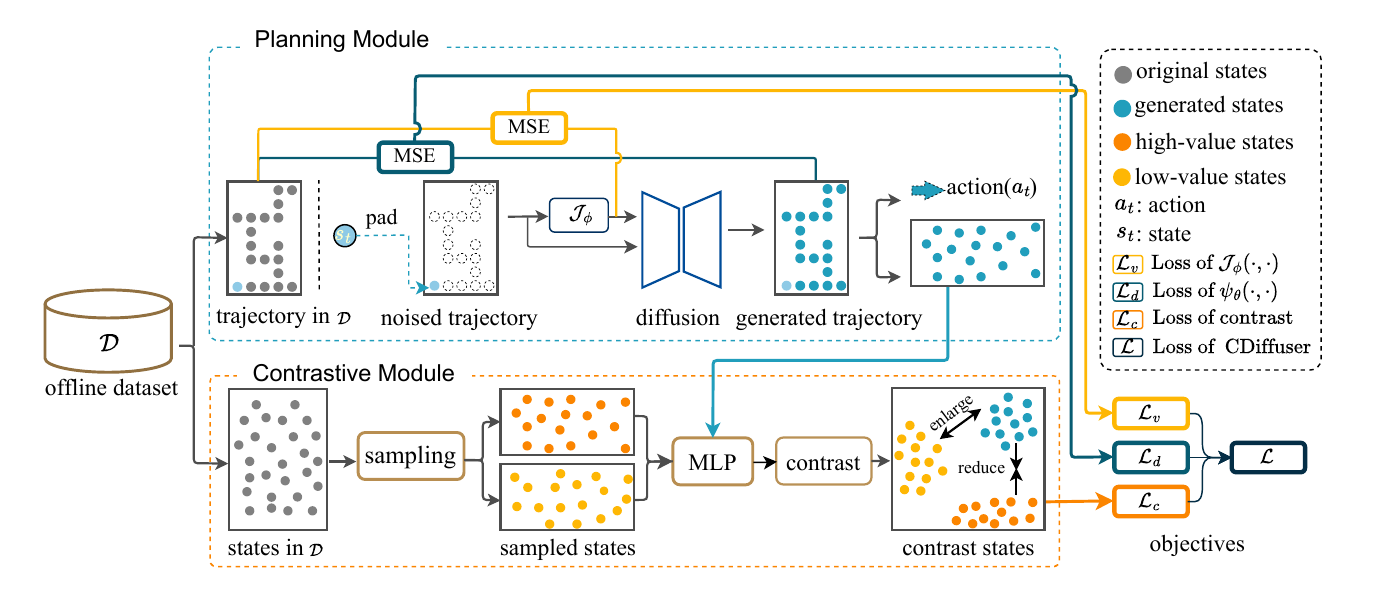}
    \vspace{-10pt}
    \caption{The overall framework of CDiffuser. 
    CDiffuser is composed of two modules: the Planning Module and the Contrastive Module. The Planning Module is designed to generate the subsequent trajectories, and the
    Contrastive Module is designed to pull the states in the generated trajectories toward the high-return states and push them away from the low-return states during the training phase.
    }
    \label{fig:framework}
    \vspace{-10pt}
\end{figure*}

As we discussed previously, the performance of offline RL methods is suppressed when the number of high-return trajectories is limited. To address the challenge, we propose a method called \textbf{Constrastive Diffuser (CDiffuser)}, which introduces a contrastive mechanism to make full use of low-return trajectories and enhance the performance by constraining the states of the agent's trajectory towards high-return states and away from low-return states. As is illustrated in \cref{fig:framework},
Our CDiffuser is composed of two modules: (1) the Planning Module, which aims to generate subsequent trajectories; (2) the Contrastive Module, which is designed to constrain the states in generated trajectories within the high-return areas and away from low-return areas.

% \shanyx{
% To improve the base distribution and effectively harnessing the inherent information within the dataset distribution, we propose...
% }

% Details of these two modules will be detailed in the following parts of this section.

% shares the simlar framework as Diffuser and is responsible for predicting trajectories; (2) the Contrastive Module, which is designed to push the learning of the base distribution towards high-return states via contrastive learning, as is illustrated in \cref{fig:framework}. Details of these two modules will be detailed in the following parts of this section.

% \zhengbang{CDiffuser is not well motivated in Methodology section. It seems a bit stiff to go straight into a detailed description of our method.}

\label{sec:Methods}
\vspace{-6pt}
\subsection{Planning Module}
\vspace{-6pt}
\label{sec:base model}
% Following \citep{janner2022planning}, given state $\bm{s}_t$ at step $t$, the Planning Module estimates the expected return $y_t$ as guidance, and leverage the guidance and the state $\textbf{s}_t$ as condition to generate the subsequent trajectory which contains the action $\textbf{a}_t$ responses to $\textbf{s}_t$, as is illustrated in Figure \ref{fig:framework}.

% Therefore, to obtain the $y_t$ for guidance, the Planning Module first pads $\textbf{s}_t$ with Gaussian noise as the initial subsequent trajectory:

Given a state $\bm{s}_t$ at step $t$, the Planning Module first generates $H$-length subsequent trajectory $\bm{\hat{\tau}}^0_t$ by alternately denoising generated trajectories and estimating trajectory returns,
% Given a state $\bm{s}_t$ at step $t$, the Planning Module estimates return $v_t$ as guidance, leverages the guidance as well as $\bm{s}_t$ as the condition to generate the $H$-length subsequent trajectory $\bm{\hat{\tau}}^0_t$, 
and then extract the action to be executed from $\bm{\hat{\tau}}^0_t$, as is illustrated in Figure \ref{fig:framework}.
Specifically, we first sample $\hat{\tau}^N_t$ from $\mathcal{N}(\bm{0},\bm{I})$, and replace $\hat{\bm{s}}_t^N$ with $\bm{s}_t$ as condition on the current observation:
% \begin{equation}
%  y_t = \text{value}(\tau^i),
% \end{equation}
\begin{equation}
 \hat{\bm{\tau}}^N_t = \{(\bm{s}_t, \hat{\bm{a}}_t^N), (\hat{\bm{s}}_{t+1}^N, \hat{\bm{a}}_{t+1}^N), ..., (\hat{\bm{s}}_{t+H}^N, \hat{\bm{a}}_{t+H}^N)\}~,
\end{equation}
in which all the elements except $\bm{s}_t$ are pure Gaussian noise. We further feed $\hat{\bm{\tau}}^N_t$ into the backward process of diffusion to generate the subsequent trajectory:
% the expected return $y_t$ and the 
% initial future trajectory to another
% diffusion process:
% \begin{equation}
% p_{\theta}(\mathbf{\hat{\tau}}^{i-1}|\mathbf{\tau}^{i}, \textbf{s}_t)=\mathcal{N}(\mathbf{\mu}_{\theta}(\mathbf{\hat{\tau}}^i, i,\textbf{s}_t), \beta_i\mathbf{I}), 
% \label{formula:ptheta}
% \end{equation}
\begin{equation} 
    p_{\theta}(\hat{\bm{\tau}}^{i-1}_t|\hat{\bm{\tau}}^{i}_t)=\mathcal{N}({\mu}_{\theta}(\hat{\bm{\tau}}^i_t, i) + \rho\nabla \mathcal{J}_{\phi}(\hat{\bm{\tau}}_t^i, i), \beta_i\bm{I})~,  \label{eq:sample}
\end{equation}
\begin{equation}
{\mu}_{\theta}(\hat{\bm{\tau}}^i_t, i) = \frac{\sqrt{\alpha^i}(1-\bar{\alpha}^{i-1})}{1-\bar{\alpha}^{i-1}}\hat{\bm{\tau}}^i_t + \frac{\sqrt{\bar{\alpha}^{i-1}}\beta^i}{1-\bar{\alpha}^i}\bm{\hat{\tau}}^{i,0}_t~.
\label{eq:sample2}
\end{equation}
% \longting{
% \begin{equation}
% {\mathbf{J}}_{\phi}(\hat{\bm{\tau}}^i_t, i) = \frac{\sqrt{\alpha^i}(1-\bar{\alpha}^{i-1})}{1-\bar{\alpha}^{i-1}}\hat{\bm{\tau}}^i_t + \frac{\sqrt{\bar{\alpha}^{i-1}}\beta^i}{1-\bar{\alpha}^i}\bm{\hat{\tau}}^{i,0}_t~.
% \label{eq:sample2}
% \end{equation}
% }

Here $\bm{\hat{\tau}}^{i,0}_t = \psi_\theta(\hat{\bm{\tau}}^i_t,i)$ represents the $\bm{{\tau}}^{0}_t$ constructed from $\hat{\bm{\tau}}_t^i$ at diffusion step $i$, $\psi_\theta(\cdot,\cdot)$ is a network for trajectory generation, $i \sim [1,N]$ is the diffusion step, $\rho$ represents the guidance scale, $\mathcal{J}_{\phi}(\cdot, \cdot)$ is a learned function to predict the return given any noisy trajectory $\bm{\tau}_t^i$.
% which is computed by function $\varphi(\cdot)$:
% \begin{equation} 
%     {v}_t^i = \varphi(\hat{\bm{\tau}}^i_t). \label{eq:pred:y}
%     % p_{\theta}(\mathbf{\tau}^{i-1}|\mathbf{\tau}^{i}, s_t)=\mathcal{N}(\mathbf{\mu}_{\theta}(\mathbf{\tau}^i, i,s_t) + \nabla y(\tau^i,i), \beta_i\mathbf{I}),  \label{eq:sample}
% \end{equation}
% where $\hat{\tau}^0_t$ is obtained by repeating \cref{eq:sample} $N$ times, and it is the future trajectory we use for planning. 
We abbreviate $\hat{\bm{\tau}}^0_t$ to $\hat{\bm{\tau}}_t$ for convenience, $\hat{\bm{\tau}}_t = \{(\bm{s}_t, \hat{\bm{a}}_t), (\hat{\bm{s}}_{t+1}, \hat{\bm{a}}_{t+1}), ..., (\hat{\bm{s}}_{t+H}, \hat{\bm{a}}_{t+H})\}$. $\hat{\bm{\tau}}_t$ is considered as \textbf{the subsequent trajectory} of $\bm{s}_t$. We take out the $\hat{\bm{a}}_t$ in $\hat{\bm{\tau}}$ as the action corresponding to the state $\bm{s}_t$. 

\vspace{-10pt}
\subsection{Contrastive Module}
% \subsection{Towards High Reward States via Contrastive Learning}
\label{sec:contrastive}
Although the Planning Module can independently generate the action responding to the environment, its performance is suppressed when the number of high-return trajectories is limited.
% due to 
% its plain base distribution and its
% neglecting of the differences among training samples.
To make use of low-return trajectories, we propose a contrastive mechanism to improve the performance by constraining the states in a subsequent trajectory toward the high-return states and away from the low-return states.
% Fortunately, this can be improved via the Contrastive Module, which makes use of low-return trajectories and applies a constrastive mechanism to constrains the states in a subsequent trajectory toward the high-return states and away from the low-return states.
% Different from the previous works~\citep{laskin2020curl,qiu2022contrastive,yuan2022robust,agarwal2020contrastive} which apply contrastive learning to obtain a better representation, we contrast the return of states for reaching high-return states. 
In the following parts, we first introduce the construction of contrastive sample sets (\textit{i.e.}, sampling the positive and negative samples for contrasting), and then we explain how we perform the contrastive mechanism.
% to improve the base distribution of the Planning Module.

\subsubsection{Sample Positive and Negative States}

\label{sec:posi_nega}

The positive samples and negative samples are necessary before applying contrastive mechanism. For an arbitrary state $\bm{s}_i \in \mathcal{S}$ in the offline dataset, we compute its return $v_i$ in advance.
Then, we propose two strategies to sample its positive sets and negative sets:

\textbf{Sampling according to return (SR).} For an arbitrary state $\bm{s}_t$ in the trajectory $\hat{\tau}_t$ generated by the Planning Module, we apply the theory of \citet{thoma2020soft} to compute the possibility of an arbitrary state $\bm{s}_i \in \mathcal{S}$ in the offline dataset is sampled as the positive sample and negative sample of state $\bm{s}_t$:
\begin{equation}
    p^+_{\bm{s}_t}(v_i) = \frac{1}{1+e^{\sigma(\xi-v_i)}}, \label{eq:posi:hard}
\end{equation}
\vspace{-5pt}
\begin{equation}
    p^-_{\bm{s}_t}(v_i) = \frac{1}{1+e^{\sigma(v_i-\zeta)}}, \label{eq:nega:hard}
\end{equation}
where $v_i$ denotes the return of $\bm{s}_i$. $p^+_{\bm{s}_t}(v_i)$ and $p^-_{\bm{s}_t}(v_i)$ denotes the probability of $\bm{s}_i$ being grouped into positive sample and negative sample of $\bm{s}_t$, correspondingly. $\xi$ and $\zeta$ are the hyper-parameters extended  from \citet{thoma2020soft}. 

%\shanyx{these parameters are not from  \citet{thoma2020soft}}
% Following \citet{thoma2020soft}, we sample the 

\textbf{Sampling according to return and dynamic consistency (SRD).} 
%\shanyx{While it is straightforward to employ the strategy of \textit{sampling according to return}, the process of pulling states towards high-return states while distancing them from low-return ones overlooks the aspect of dynamic consistency.}
Though the strategy of \textit{sampling according to return} is easy to deploy, pulling the states toward high-return states and away from low-return states neglects the dynamic consistency. 
Hence, for the states in the offline dataset, we additionally conduct MiniBatch K-Means clustering ~\citep{sculley2010web}, and compute the transition probability of clusters, \textit{i.e.}, the frequency of the states in one cluster transit to another cluster. For an arbitrary state $\bm{s}_t$ in the trajectory $\hat{\tau}_t$ generated by the Planning Module, we compute its cluster by K-Means, and obtain the candidate set $\mathcal{S}_t$ of the subsequent states according to the transition probability among clusters. Then for $\bm{s}_i \in \mathcal{S}_t$, we sample the positive sample of $\bm{s}_t$ by Eq.~\ref{eq:posi:hard}.  For $\bm{s}_i \in \mathcal{S}$ in offline dataset, we sample the negative sample of $\bm{s}_t$ by Eq.~\ref{eq:nega:hard}.

\subsubsection{Constrain the trajectory with contrastive learning}
% As is discussed in \cref{sec:base model}, the generated trajectory is $\hat{\bm{\tau}} = \{(\bm{s}_t, \hat{\bm{a}}_t), (\hat{\bm{s}}_{t+1}, \hat{\bm{a}}_{t+1}), ..., (\hat{\bm{s}}_{t+H}, \hat{\bm{a}}_{t+H})\}$. 
% Following ~\citet{kang2023efficient}, i
% Instead of running the whole reverse denoising process to sample $\hat{\tau}_t$ for contrastive, we cheaply contruct $\bm{\hat{\tau}}^{i,0}_t = \{(\hat{\bm{s}}^{i,0}_t, \hat{\bm{a}}^{i,0}_t), (\hat{\bm{s}}^{i,0}_{t+1}, \hat{\bm{a}}^{i,0}_{t+1}), ..., (\hat{\bm{s}}^{i,0}_{t+H}, \hat{\bm{a}}^{i,0}_{t+H})\}$ from $\bm{\tau}^i_t$ by performing one-step denoising.

To constrain the states in subsequent trajectories while avoiding the cost of running the whole backward denoising process, we leverage the noised trajectory in the diffusion backward process to reconstruct a neat trajectory, \ie, $\bm{\hat{\tau}}^{i,0}_t = \{(\hat{\bm{s}}^{i,0}_t, \hat{\bm{a}}^{i,0}_t), (\hat{\bm{s}}^{i,0}_{t+1}, \hat{\bm{a}}^{i,0}_{t+1}), ..., (\hat{\bm{s}}^{i,0}_{t+H}, \hat{\bm{a}}^{i,0}_{t+H})\}$ from $\bm{\tau}^i_t$ for any arbitrary diffusion step $i$.
% \shanyx{As is discussed in \cref{sec:base model}, the Planning Module denoises $\bm{\tau}^i_t$ to $\bm{\tilde{\tau}}^{0}_t = \{(\tilde{\bm{s}}^0_t, \tilde{\bm{a}}^0_t), (\tilde{\bm{s}}^0_{t+1}, \tilde{\bm{a}}^0_{t+1}), ..., (\tilde{\bm{s}}^0_{t+H}, \tilde{\bm{a}}^0_{t+H})\}$ during the diffusion step $i$.
% }
% To constrain the states in this trajectory, 
Then, we extract states in $\hat{\bm{\tau}}_t^{i,0}$ as $\mathcal{S}_{\hat{\bm{\tau}}^{i,0}_t} = \{\hat{\bm{s}}_{t+1}^{i,0},\hat{\bm{s}}_{t+2}^{i,0},...,\hat{\bm{s}}_{t+H}^{i,0}\}$.
% We will introduce the auxiliary objective based on contrastive learning in this section. 
% For the sake of discussion, let's assume that we are training CDiffuser with sample $\tau=\{(s_t, a_t), (s_{t+1}, a_{t+1}), ..., (s_{t+H}, a_{t+H})\}$. At each training step, Trajectory Generation Module outputs a reconstructed version, $\hat{\tau} = \{(s_t, \hat{a}_t), (\hat{s}_{t+1}, \hat{a}_{t+1}), ..., (\hat{s}_{t+H}, \hat{a}_{t+H})\}$ based on $\tau^i$. To perform contrastive learning on $\hat{\tau}$, firstly, we extract states in $\hat{\tau}$ as $\{\hat{s}_{t+1},\hat{s}_{t+2},...,\hat{s}_{t+H}\}$. 
% For each state $\hat{\bm{s}}_h^{i,0} \in \mathcal{S}_{\hat{\bm{\tau}}^{i,0}_t}$, we sample $\kappa$ states via \cref{eq:possible:posi} as positive samples and $\kappa$ states via \cref{eq:possible:nega} as negative samples from the offline dataset. %These samples form a positive sample set $\mathcal{S}_h^{+}$ and a negative sample set $\mathcal{S}_h^{-}$ respectively.
For each state $\hat{\bm{s}}_h^{i,0} \in \mathcal{S}_{\hat{\bm{\tau}}^{i,0}_t}$, we sample $\kappa$ states as positive sample set $\mathcal{S}_h^{+}$  and $\kappa$ states as negative sample set $\mathcal{S}_h^{-}$ from the offline dataset.

Inspired by \citep{schroff2015facenet} and \citep{sohn2016improved}, to apply contrastive learning to the scenario of multiple positive samples and impose aggressive constraints, we removed the positive sample term from the denominator polynomial in \cref{eq:infonce} and propose the following equation to pull the states in the generated subsequent trajectory toward the high-return states and away from the low-return states: 
% To apply \cref{eq:infonce} to multiple positive samples scenario and impose aggressive constraints, inspired by \citet{schroff2015facenet} 
%  and \citet{sohn2016improved}, we removed the positive sample terms from the denominator polynomial and propose the following equation to pull the states in the generated trajectory toward the high-return states and away from the low-return states:
% \begin{equation}
%         L=-\log\mathbb{E}_{\textbf{s}^{+} \in\mathcal{S}_j^{+}, \textbf{s}^{-}\in\mathcal{S}_j^{-}}\left[\frac{\text{exp}( \text{sim}(\hat{\textbf{s}}_j, \textbf{s}^{+}) / \mathcal{T} )}{  \sum \text{exp}( \text{sim}(\mathbf{x}, \textbf{s}^{-}) / \mathcal{T})  }\right],  \label{eq:infonce}
% \end{equation}
% Following Eq.(\ref{{eq:infonce}})
% % \cref{eq:infonce}, 
% the objective of each state is then formulated as:
\begin{equation}
% \vspace{-10pt}
    % \mathcal{L}_s = \frac{1}{b}\sum_{i=0}^bd(\hat{s}, \mathcal{S})
    \mathcal{L}_{h}^i =-\log \frac{ \sum_{k=0}^{\kappa} \exp( \text{sim}( {f}(\hat{\bm{s}}_h^{i,0}),  {f}({\bm{s}}_h^{+})  ) / T ) }{ \sum_{k=0}^{\kappa} \exp( \text{sim}( {f}(\hat{\bm{s}}_h^{i,0}), {f}({\bm{s}}_h^{-}) ) / T )  }~, \label{eq:loss:contrast:sub}
\end{equation}
where $\bm{s}_h^{+} \in \mathcal{S}_h^{+}$, $\bm{s}_h^{-} \in \mathcal{S}_h^{-}$. ${f}(\cdot)$ represents the projection function, $T$ represents the temperature~\citep{wang2021understanding}, and $\text{sim}(\cdot,\cdot)$ denotes the cosine similarity, which is computed as
\begin{equation}
    \text{sim}(\bm{a}, \bm{b}) = \frac{\bm{a}^\top\bm{b}}{\|\bm{a}\|\cdot\|\bm{b}\|}~.
\end{equation}

% \citep{oord2018representation}
It is worth noting that Eq.(~\ref{eq:loss:contrast:sub}) is different from the standard infoNCE loss \citep{oord2018representation}, we made these modifications primarily for the sake of the model's effectiveness.

\subsection{Model Learning}
\label{sec:model_learn}
% Since the final action is one of the elements in the generated trajectory, and it is influenced by $y_t$ and constrained by contrastive learning. \shanyx{
Recall that the action responding to state $\bm{s}_t$ is one of the elements in the generated trajectory, which is influenced by the return predictor $\mathcal{J}_\phi(\cdot, \cdot)$ and constrained by contrastive learning. Therefore, we optimize our method from the perspective of trajectory generation, return prediction, and contrastive learning constrain. 

Specifically, we optimize the trajectory generation by minimizing the Mean Square Error between the ground truth and neat trajectory predicted by $\psi_\theta(\cdot,\cdot)$ given any intermediate noisy trajectories as input:
\begin{equation}
    \mathcal{L}_d = \mathbb{E}_{\bm{\tau}_t\in \mathcal{D}, t>0, i \sim [1, N]}\left[\| \bm{\tau}_t - \psi_\theta(\bm{\tau}^i_t,i) \|^2\right]~, \label{eq:loss:difusion}
\end{equation}
where $i$ denotes the step of diffusion, $\bm{\tau}^i_t$ is obtained in the $i$-th step of forward process.

We optimize the return predictor by minimizing the Mean Square Error between the predicted return $\mathcal{J}_\phi(\bm{\tau}_t^i, i)$ and the ground-truth return $v_t$:
\begin{equation}
    % \mathcal{L}_v = \text{MSE}(\varphi(\bm{\tau_t^i}), v_t),
    \mathcal{L}_v = \mathbb{E}_{\bm{\tau}_t\in \mathcal{D}, t>0, i \sim [1, N]}[\| \mathcal{J}_\phi(\bm{\tau}_t^i, i) - v_t\|^2]~.
\label{eq:loss:y}
\end{equation}

% The objective function of our conditioned diffusion model is then given by:

We constrain the trajectory generation with a weighted contrastive loss:
\begin{equation}
    \mathcal{L}_c =
    \mathbb{E}_{t>0, i \sim [1, N]}\left[\sum_{h=t}^{t+H}\frac{1}{h+1}\mathcal{L}_{h}^i\right],
    \label{eq:loss:contrast:all}
\end{equation}
in which the coefficient $\frac{1}{h+1}$ decreases as $h$ increases since the importance gradually diminishes as it approaches the end of the planning horizon.
% we design a coefficient sequence that follows an $y=x^{-1}$ decay to reweight the contrastive loss within the trajectory.

% \begin{equation}
%     \mathcal{L}_d=\mathbb{E}_{\mathbf{\tau}_0, i \sim \{1,2,...,N\}  }[|| \tau^0 - \tau_\theta(\tau^i,i) ||^2],\label{eq:loss:difusion}
% \end{equation}
%  in which $i \sim \mathcal{U}\{1,2,...,N\}$ is the diffusion step, $\tau_\theta$ is a model to recontruct $\tau^0$. 

% Recall that CDiffuser predicts future $H$ steps trajectory beyond the current time step. As a result, we need to carefully design the reduction of contrastive loss for each state within the trajectory, rather than naively summing or averaging it.

% Taking into account that the model's prediction confidence tends to decrease as $H$ increases, we design a coefficient sequence that follows an $y=x^{-1}$ decay to reweight the contrastive loss within the trajectory. We then have the objective for contrastive learning $\mathcal{L}_c$:

% \begin{equation}
%     \mathcal{L}_c = \sum_{h=0}^{H}\frac{1}{h+1}\mathcal{L}_{\hat{s}_{t+h}}.\label{eq:loss:contrast:all}
% \end{equation}

Hence, the overall objective function of CDiffuser can be written as a weighted sum of the aforementioned loss terms:
\begin{equation}
    \mathcal{L} = \lambda_d \mathcal{L}_d +  \lambda_v \mathcal{L}_v + \lambda_c \mathcal{L}_c~,
    \label{eq:loss:all}
\end{equation}
where $\lambda_d$, $\lambda_v$, $\lambda_c$ are hyperparameters, which balance the importances of the corresponding learning targets. Please note that the return predictor $\mathcal{J}_\phi(\cdot, \cdot)$ and $\psi_\theta(\cdot,\cdot)$ are independent, thus optimizing $\mathcal{J}_\phi(\cdot, \cdot)$ and $\psi_\theta(\cdot,\cdot)$ with $\mathcal{L}$ is identical to separately optimizing $\mathcal{J}_\phi(\cdot, \cdot)$ with $\mathcal{L}_v$ and $\psi_\theta(\cdot,\cdot)$ with $\mathcal{L}_d$ and $\mathcal{L}_c$. Please refer to the proof in \cref{appendix:sec:loss} for details. 

The pseudo code of CDiffuser is presented in \cref{appendix:psuedo:all},  and the details of implementation will be discussed in the next section.

\section{Experiments}
\label{sec:exp}

In this section, we evaluate the performance of CDiffuser across a wide variety of tasks. We first demonstrate the advantages of CDiffuser on 14 standard D4RL datasets. 
%Next, the results of CDiffuser and baselines on manually constructed high variance datasets demonstrate the significant advantage of CDiffuser in scenarios with a small number of expert samples, showcasing its ability to extract expert information from low-quality datasets. 
Next, we construct high variance datasets and evaluate the performance of CDiffuser and baselines on them. The results demonstrate the significant advantage of CDiffuser in scenarios with low-quality datasets, showcasing CDiffuser's ability to extract expert information.
Further, we delve into more comprehensive experiments to analysis the key designs of CDiffuser as well as the portability of CDiffuser.

% The overall results are detailed in \cref{sec:overall result}. Further, we conduct experimental analyses on the designs and key hyperparameters in our model respectively, in \cref{sec:ablation} and \cref{sec:hyper}. We also perform further investigations in \cref{sec:further investigation}.

% For simplicity, \textbf{CDiffuser} is denoted as \textbf{CDiffuser} in tables and figures of this section. \shanyx{mark}

\subsection{Experiment Settings}
\label{sec:setting}
\textbf{Environments and datasets.}
We evaluate the performance of CDiffuser on the locomotion tasks, navigation tasks and manipulation tasks. Specifically, we evaluate the locomotion capability of CDiffuser on Halfcheetah, Hopper, Walker2d, evaluate the navigation capability of CDiffuser on Maze2d, and evaluate the ability of CDiffuser in complex tasks on Kitchen. For each environment, we train CDiffuser with various scales of offline datasets provided by D4RL~\citep{fu2020d4rl}, and test the performance of CDiffuser on the corresponding environments.

% These environments simulate classic physical systems that require continuous control, given the environment states, the models are supposed to determine the angles for each joint that the agent should take as its actions. The final target is to train a policy so that it can control the agent to move as far as possible within a limited number of steps, essentially maximizing the discounted accumulated reward.

% CDRL and baselines are trained on offline datasets provided by D4RL containing trajectories collected by different polices, $\ie$, med-expert, med and med-replay. Nevertheless, we evaluate the models trained on different datasets in a unified environment.

\textbf{Baselines.} 
We compare CDiffuser with diffusion-free methods such as CQL~\citep{kumar2020conservative}, IQL~\citep{kostrikov2021offline}, MOPO~\citep{yu2020mopo}, Decision Transformer (DT)~\citep{chen2021decision} and Trajectory Transformer (TT)~\citep{janner2021offline}. Further, we compare CDiffuser with diffusion-based methods  Diffuser~\citep{janner2022planning} and Decision Diffuser (DD)~\citep{ajay2023is}, which apply diffusion to model RL as sequence generation problems.
% share the same fundamental structure as CDiffuser.
%We compare our model with value-based offline RL algorithms, CQL~\citep{kumar2020conservative} and IQL~\citep{kostrikov2021offline}, and a transformer-based sequence modeling algorithm, DT~\citep{chen2021decision}. Further, we compare CDiffuser with state-conditioned trajectory generative models Diffuser~\citep{janner2022planning} and Decision Diffuser~\citep{ajay2023is}, which share the same fundamental structure as CDiffuser.

% We report the performance of baseline methods using the best results reported from their own paper. 

\textbf{Implementation details.} 
\label{sec:resource} We adopt U-Net~\citep{ronneberger2015u} as the denoise network $\psi_\theta(\cdot,\cdot)$ and the return predictor $\mathcal{J}_\phi(\cdot, \cdot)$, and adopt a linear layer with $Sigmoid$ as the activation function as the projector $f(\cdot)$. Our model is trained on a device with 4 NVIDIA A40 GPUs, Intel Gold 5220 CPU and 504G memory, optimized by Adam~\citep{kingma2014adam} optimizer.

\vspace{-10pt}
\subsection{Benchmark Results}
\label{sec:benchmark}
\vspace{-6pt}
\label{sec:overall result}
We compare CDiffuser to baseline methods with respect to the normalized average returns~\citep{fu2020d4rl} obtained during online evaluation.
We conducted 10 trials with different seeds and reported the average results. The results of CDiffuser and baseline methods are summarized in \cref{tab:data}, in which \textbf{CDiffuser-SR} denotes the states used for contrasting are sampled with SR strategy and \textbf{CDiffuser-SRD} denotes the states used for contrasting are sampled with SRD strategy.

\renewcommand\arraystretch{1.1}
\begin{table*}[t!]
\vspace{-10pt}
    % \vspace{-15pt}

    % \caption{Offline Reinforcement Learning Performance. The performance of baseline methods are reported using the same results reported in their own papers.  Optimal and sub-optimal results of each setting are marked as \textbf{bold} and \underline{underline}, respectively.}
    \caption{The average normalized score of different methods on various environments, with $\pm$ denoting the standard deviation. The mean and standard deviation are computed over 50 random seeds. The best and the second-best results of each setting are marked as \textbf{bold} and \underline{underline}, respectively. 
    % The shaded column represents our method.
    }
    \vspace{5pt}
    \centering
    % \vspace{-5pt}
    % \scriptsize
    % \begin{lrbox}{\tablebox}
    % \hspace{-5pt}
    % \begin{tabular}{p{1.8cm}|r|c|l|l|l|l|l|l|l|l}
    \resizebox{1\linewidth}{!}{
    \begin{tabular}{l l c c c c c | c c | c c}
        \toprule
         \textbf{Dataset} & \textbf{Environment}    & \textbf{CQL}      & \textbf{IQL}  & \textbf{DT}   & \textbf{TT}  &\textbf{MOPO}  & \textbf{Diffuser} & \textbf{DD} & \textbf{CDiffuser-SR} & \textbf{CDiffuser-SRD} \\
        \midrule
        Med-Expert & HalfCheetah                    & {91.6}            & 86.7          & 86.8          & \textbf{95.0} & 63.3   & 88.9 $\pm$ 0.3 & 90.6 $\pm$ 1.3 & \underline{92.0 $\pm$ 0.4}  &  89.9  $\pm$ 0.6  \\
        Med-Expert & Hopper                         & 105.4             & 91.5          & 107.6         & 110.0         & 23.7 & 103.3 $\pm$ 1.3 & \underline{111.8} $\pm$ 1.8 & \textbf{112.4 $\pm$ 1.2 }  &  106.4  $\pm$ 1.3  \\
        Med-Expert & Walker2d                       & \underline{108.8} & \textbf{109.6}& 108.1         & 101.9         & 44.6 & 106.9 $\pm$ 0.2 & \underline{108.8} $\pm$ 1.7 &  108.2 $\pm$ 0.4  &  106.7  $\pm$ 2.2  \\
        \midrule
        Medium & HalfCheetah                        & 44.0              &\underline{47.4}& 42.6         & {46.9}        & 42.3& 42.8 $\pm$ 0.3 & \textbf{49.1} $\pm$ 1.0 &  43.9 $\pm$ 0.9  &  42.9  $\pm$ 0.2     \\
        Medium & Hopper                             & 58.5              & 66.3          & 67.6          & 61.1          & 28.0  & 74.3 $\pm$ 1.4 & \underline{79.3} $\pm$ 3.6 & \textbf{92.3 $\pm$ 2.6} & 75.2  $\pm$ 1.1    \\
        Medium & Walker2d                           & 72.5              & 78.3          & 74.0          & 79.0          & 17.8 & 79.6 $\pm$ 0.55 & \underline{82.5} $\pm$ 1.4 &  \textbf{82.9 $\pm$ 0.5}  & 82.2  $\pm$ 1.1   \\
        \midrule
        Med-Replay & HalfCheetah                    & \underline{45.5}     &44.2 & 36.6         & 41.9          & \textbf{53.1}  & 37.7 $\pm$ 0.5 & 39.3 $\pm$ 4.1 & 40.0 $\pm$ 1.1  & 36.6  $\pm$ 2.9  \\
        Med-Replay & Hopper                         & 95                & 94.7          & 82.7          & 91.5          & 67.5 & 93.6 $\pm$ 0.4 & \textbf{100} $\pm$ 0.7  &  \underline{96.4 $\pm$ 1.1}  &  95.5  $\pm$ 0.9  \\
        Med-Replay & Walker2d                       & 77.2              & 73.9          & 66.6          & \underline{82.6} & 39.0       & 70.6 $\pm$  1.6 & 75 $\pm$ 4.3 & \textbf{84.2 $\pm$ 1.2}  &  75.3  $\pm$ 1.7 \\
        \midrule

        % \hline
        U-Maze  & Maze2d   & 5.7  & 47.4  & 9.2 & 25.4 & 13.6  & \underline{113.9} $\pm$ 3.1 & 0.0 &  \textbf{142.9}  $\pm$ 2.2  &  85.6  $\pm$ 3.2   \\
        Medium  & Maze2d   & 5.0  & 34.9  & 9.6 & 23.3 & 33.3  & \underline{121.5} $\pm$ 2.7 & 0.0 &  \textbf{140.0}  $\pm$ 0.7  &  115.7  $\pm$ 0.7  \\
        Large   & Maze2d   & 12.5 & 58.6  & 10.4 & 27.7 & 0.0  & \underline{123.0} $\pm$ 6.4 & 0.0 &  \textbf{131.5}  $\pm$ 3.2  &  99.7   $\pm$ 1.2   \\
        \midrule
        
        % \hline
        Mixed   & Kitchen   & 52.4  & 51.0  & 20.9 & 31.1 & 0.0  & 42.5  $\pm$ 1.9  & \underline{65.0} $\pm$ 2.8 &  \textbf{65.0}  $\pm$ 1.3 &  32.5 $\pm$ 2.2 \\
        Partial & Kitchen   & 51.2  & 46.3  & 35.2 & 32.9 & 0.0  & 40.0  $\pm$ 3.1  & \underline{57.0} $\pm$ 2.5 &  \textbf{58.0}  $\pm$ 1.9  &  40.0 $\pm$ 3.1 \\
        \bottomrule
        % \hline
        
    \end{tabular}
    }
    % \end{lrbox}
    % \scalebox{1.2}{\usebox{\tablebox}}
    % \vspace{5pt}
    \label{tab:data}
\end{table*}

From Table \ref{tab:data}, we can observe that: (1) Compared with all the baseline methods, CDiffuser achieves the best or the second-best performance on 6 out of 9 locomotion tasks (HalfCheetah, Hopper, and Walker2d) and achieves the best performance on all the two high-dimensional manipulation tasks (Kitchen), demonstrating the outstanding performance of CDiffuser under periodic settings.
Moreover, CDiffuser achieves the best performance on all the three navigation tasks, demonstrating the excellent ability of CDiffuser in long-term planning. 
% When it comes to the Kitchen environment, CDiffuser outperforms all the baselines on 2 out of 2 tasks, demonstrating the ability of CDiffuser in complex tasks.
% (2) Compared with our backbone method Diffuser, CDiffuser outperforms Diffuser in all the 14 tasks, which demonstrates the effectiveness of contrast in boosting diffusion-based RL methods. 
(2) Compared to the methods with similar backbones, Diffuser outperforms Diffuser in all 14 tasks, and outperforms DD in 11 tasks, which demonstrates the benefits of introducing the constrastive mechanism. 
We can observe that CDiffuser exhibits more improvement in Med and Med-Replay datasets than the Med-Expert datasets. Since Med-Expert datasets have more high-return samples, they offer abundant information for methods like Diffuser to learn, thus they can achieve better results. However, both Med and Med-Replay have more low-return samples than Med-Expert, which increases the difficulty of learning a good policy. These results demonstrate that CDiffuser is more effective on datasets with many low-return trajectories.

% \subsection{Performance with Poor-expert Datasets}
\subsection{Study on the limited number of high-return trajectories.}
\renewcommand\arraystretch{1.1}
\begin{table*}[t!]
    % \vspace{-15pt}

    % \caption{Offline Reinforcement Learning Performance. The performance of baseline methods are reported using the same results reported in their own papers.  Optimal and sub-optimal results of each setting are marked as \textbf{bold} and \underline{underline}, respectively.}
    \caption{The average normalized score of CDiffuser and baseline methods on various datasets of Halfcheetah mixed with different ratios of expert data, with $\pm$ denoting the standard deviation. We compute the mean and standard deviation over 50 random seeds for CDiffuser, and 10 random seeds for baselines. The best and the second-best results of each setting are marked as \textbf{bold} and \underline{underline}, respectively. Ratio denotes the the ratio of trajectories from the Expert dataset.  
    }
    \vspace{5pt}
    \centering
    % \vspace{-5pt}
    % \scriptsize
    % \begin{lrbox}{\tablebox}
    % \hspace{-5pt}
    % \begin{tabular}{p{1.8cm}|r|c|l|l|l|l|l|l|l|l}
    \resizebox{1\linewidth}{!}{
        \begin{tabular}{@{}l|c|ccccc|cc|cc@{}}
            \toprule
            \multicolumn{1}{l|}{\textbf{Dataset}} & \textbf{Ratio} & \textbf{CQL} & \textbf{IQL}  & \textbf{DT} & \textbf{TT} & \textbf{MOPO} & \textbf{Diffuser} & \textbf{DD} & \textbf{CDiffuser-SR} & \textbf{CDiffuser-SRD} \\ \midrule
            \multirow{3}{*}{M-Exp}              & 0.1              & 34.2   $\pm$ 1.1       & 51.56    $\pm$ 5.4     & 58.6    $\pm$ 2.1     & 46.7     $\pm$ 1.1    & 67.1    $\pm$ 0.3       & 71.5 $\pm$ 1.8 & 43.2   $\pm$ 0.9      & \underline{ 72.1}   $\pm$ 0.9         & \textbf{73.6}   $\pm$ 1.2      \\
                                                 & 0.2              & 39.4   $\pm$ 1.7       & 62.9     $\pm$ 0.5      & 46.8    $\pm$ 0.5     & 46.9     $\pm$ 1.8    & 27.2    $\pm$ 3.8       & \underline{ 80.3}  $\pm$ 0.8       & 41.6    $\pm$ 2.2    & \textbf{81.3}   $\pm$ 1.3      & 77.5    $\pm$ 0.9              \\
                                                 & 0.3                 & 63.1  $\pm$ 0.8        & 82.3    $\pm$ 2.8       & 70.9   $\pm$ 1.2      & 47.4    $\pm$ 1.1     & 38.7   $\pm$ 2.2        & \underline{ 81.7} $\pm$ 2.9        & 43.6    $\pm$ 3.3     & \textbf{82.8}    $\pm$ 1.2     & 68.3     $\pm$ 1.1             \\ \midrule
            \multirow{3}{*}{MR-Exp}          & 0.1       & 39.8  $\pm$ 1.4        & \underline{44.2} $\pm$ 2.2 & 7.5    $\pm$ 2.1      & 42.9  $\pm$ 2.5  & \textbf{55.7} $\pm$ 0.9         & 38.1       $\pm$ 1.1        & 33.8   $\pm$ 1.6     & 42.2 $\pm$ 0.7              & 39.0      $\pm$ 0.5           \\
                                                 & 0.2           & 37.6  $\pm$ 2.9        & 48.5    $\pm$ 1.8       & 6.7    $\pm$ 4.5      & 43.7    $\pm$ 1.7     & 40.6   $\pm$ 1.1        & 46.2              $\pm$ 0.2 & 32.8    $\pm$ 0.6    & \underline{ 50.6}   $\pm$ 1.2        & \textbf{58.4}  $\pm$ 0.9      \\
                                                 & 0.3             & 40.3  $\pm$ 0.8        & 44.6    $\pm$ 1.8       & 6.1    $\pm$ 0.1      & 49.3    $\pm$ 2.2     & 44.1   $\pm$ 0.9        & \underline{ 57.1} $\pm$ 1.8        & 36.2    $\pm$ 1.1    & \textbf{60.3}     $\pm$ 2.7   & 55.9   $\pm$ 1.5              \\ \midrule
            \multirow{3}{*}{Rand-Exp}              & 0.1           & 31.1  $\pm$ 4.4        & 3.9     $\pm$ 0.0       & 5.1    $\pm$ 0.0      & 7.7     $\pm$ 0.1     & 23.8   $\pm$ 5.7        & \underline{ 33.8} $\pm$ 1.8        & 13.8   $\pm$ 0.8      & 18.1      $\pm$ 1.1            & \textbf{48.0}  $\pm$ 2.9      \\
                                                 & 0.2          & 39.4  $\pm$ 1.9        & {72.9} $\pm$ 2.6 & 10.3   $\pm$ 4.2      & 16.8    $\pm$ 1.8     & 30.6   $\pm$ 0.8        &  \underline{74.4} $\pm$ 1.5        & 8.5  $\pm$ 0.2       & 72.3   $\pm$ 0.7              & \textbf{77.6}    $\pm$ 0.9             \\
                                                 & 0.3              & 38.4  $\pm$ 0.4        & 50.7    $\pm$ 1.1       & 27.5   $\pm$ 6.8      & 5.9     $\pm$ 0.2     & 30.6   $\pm$ 0.1        & 75.8    $\pm$ 2.3     & 13.9  $\pm$ 1.1      & \underline{ 86.6}   $\pm$ 1.8         & \textbf{88.7}     $\pm$ 0.9    \\ \bottomrule
            \end{tabular}
    }
    % \end{lrbox}
    % \scalebox{1.2}{\usebox{\tablebox}}
    % \vspace{5pt}
    \label{tab:data-mix}

\vspace{-10pt}
\end{table*}

As we discussed previously, the proportion of high return trajectories has a significant impact on the performance of the model. To investigate the performance of CDiffuser in different ratios of high-return trajectories, we mix different trajectories from Halfcheetah and obtain three datasets:
\begin{itemize}[leftmargin=10pt]

    \item \textbf{M-Exp}: mix the the trajectories in Medium and Expert.
    \item \textbf{MR-Exp}: mix the the trajectories in Med-Replay and Expert. 
    \item \textbf{Rand-Exp} mix the trajectories in Expert and the trajectories sampled by random policy interacting with environment.
\end{itemize}
We set the ratio of trajectories from the Expert dataset to 0.1, 0.2, and 0.3, resulting in three different settings of these datasets. The visualization of returns is illustrated in Figure \ref{fig:mixed_halfcheetah}.

We test the performance of the baselines and CDiffuser on the three datasets, and the results are illustrated in \cref{tab:data-mix}. From the results, we can observe that:  (1) In most cases, offline RL methods' performance declines with the ratio declines, in which a small ratio indicates fewer high-return trajectories. Moreover, compared with the performance in Med-Expert, most baselines have a performance decline when trained with Rand-Exp, which has fewer high-return trajectories than Med-Expert. These results validate that the performance of offline RL methods is influenced by the proportion of high-return trajectories in the dataset; (2) CDiffuser demonstrates more significant improvement in Rand-Exp than M-Exp and MR-Exp, for instance, it outperforms the best baseline (Diffuser) by 1.1 in M-Exp (ratio = 0.3), but it outperforms the best baseline (Diffuser) by 12.9 in Rand-Exp (ratio = 0.3). That demonstrates CDiffuser has significant superiority in the case with limited number of high-return trajectories.

\vspace{-10pt}
\subsection{Further Investigation}
\label{sec:further investigation}
% \vspace{-10pt}

% \begin{figure*}[t]
% \vspace{-6pt}
% \centering
% \includegraphics[width=0.9\linewidth]{figs/heatmap_ICML_-0.pdf} 
% \vspace{-10pt}
% \caption{The similarities between the states in the generated trajectories and actual states. The generated states of CDiffuser are more similar with the actual states, demonstrating the better long-term dynamic consistency.
% }
% \vspace{-15pt}
% \label{fig: visual2}
% \end{figure*}

% \begin{figure}[t]
% \centering
% \includegraphics[width=1.05\columnwidth]{figs/heatmap_ICML_sc-0.pdf} 
% \caption{The similarities between the states in the generated trajectories and actual states. The generated states of CDiffuser are more similar with the actual states, demonstrating the better long-term dynamic consistency.
% }
% \label{fig: visual2}
% \end{figure}

To further investigate the performance of CDiffuser, we conduct ablation study, and analyze the state-reward distribution as well as the reward distribution.
% the base distribution of CDiffuser.}
\vspace{-10pt}
\paragraph{Ablation studies.}
\label{sec:ablation}

% \begin{figure*}[ht]
%     \centering
%     % \hspace{-5pt} 
%     \subfigure[Dataset]{
%         \begin{minipage}[t]{0.34\linewidth}
%         \centering
%         \includegraphics[width=1.0\linewidth]{figs/vis_dataset_low.pdf}
%         \end{minipage}%
%     }%
%     % \hspace{10pt}
%     \subfigure[Diffuser]{
%         \begin{minipage}[t]{0.34\linewidth}
%         \centering
%         % \vspace{-130pt}
%         \includegraphics[width=1.0\linewidth]{figs/vis_diffuser_low.pdf}
%         \end{minipage}%
%     }%
%    % \hspace{10pt}
%     \subfigure[CDRL]{
%         \begin{minipage}[t]{0.34\linewidth}
%         \centering
%         % \vspace{-130pt}
%         \includegraphics[width=1.0\linewidth]{figs/vis_plan14bf_low.pdf}
%         \end{minipage}%
%     }%
%     \caption{Rewards of different states in dataset, and of states collected by Diffuser and CDRL.}
%     \label{fig:generated state-reward distribution}
% \end{figure*}

% \subsubsection{Settings}
We have the following variants to conduct ablation study:
\vspace{-5pt}
% \vspace{-10pt}
\begin{itemize}[leftmargin=10pt]

    \item \textbf{CDiffuser-N}: only apply the samples with high-return to train the model.
    
    \item \textbf{CDiffuser-C}: remove contrastive mechanism from CDiffuser, \ie, remove $\mathcal{L}_c$ from \cref{eq:loss:all}.
    \vspace{-5pt}
    % Set $\sigma=0$, \ie, remove contrastive learning from CDiffuser. CDiffuser then collaps into learning with Trajectory Generation Module only.
    
    % Remove contrastive learning from CDiffuser and train the Trajectory Generation Module with positive samples only. Please notice than the contrastive sets are obtained in the manner of CDiffuser-C.

    % \item \textbf{CDiffuser-G}: remove the guidance from CDiffuser, \ie, removing $\rho\nabla \mathcal{J}_\phi(\cdot, \cdot)$ from \cref{eq:sample}.

    % % The structure and training process are consistent with CDiffuser, but guidance is not applied during sampling. 
    % \item \textbf{Diffuser-G}: remove the classifier guidance from Diffuser.

    % The structure and training process are consistent with Diffuser, but guidance is not applied during sampling. 
\end{itemize}

\begin{figure}[t]
\centering
\includegraphics[width=1.\columnwidth]{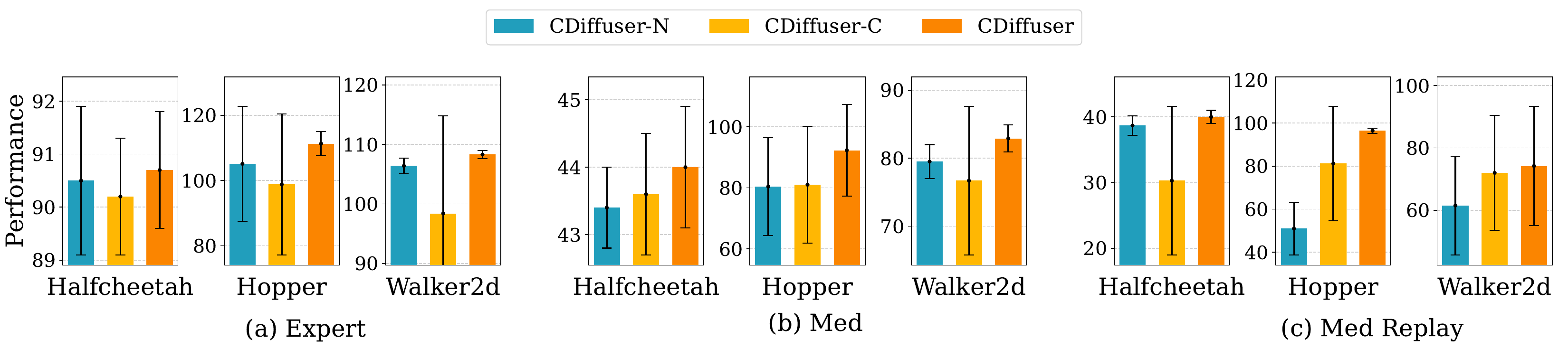} 
\caption{Results of the ablation experiments on different variants.
}
\label{fig:abla}
\end{figure}

The results are summarized in Figure \ref{fig:abla}, from which we can conclude the following key findings:

% \begin{wrapfigure}{r}{0.5\textwidth}
%     \includegraphics[width=0.48\columnwidth]{figs/abla_singc_ng.pdf} 
%     \caption{Results of the ablation experiments on different variants.    }
%     \label{fig:abla}
% \end{wrapfigure}

% \begin{wrapfigure}{l}{0.5\textwidth}
%     \includegraphics[width=0.48\columnwidth]{figs/states_value_markedingle.jpg} 
%     \caption{The distribution of state and reward. It is better to view in color mode. CDiffuser achieves higher rewards in out-of-distribution areas (circled with red).
%     }
%     \label{fig: visual1}
% \end{wrapfigure}

\begin{figure}[t]
% \vspace{-10pt}
  \centering
  \begin{minipage}[b]{0.55\textwidth}
    \centering
    \includegraphics[width=\textwidth]{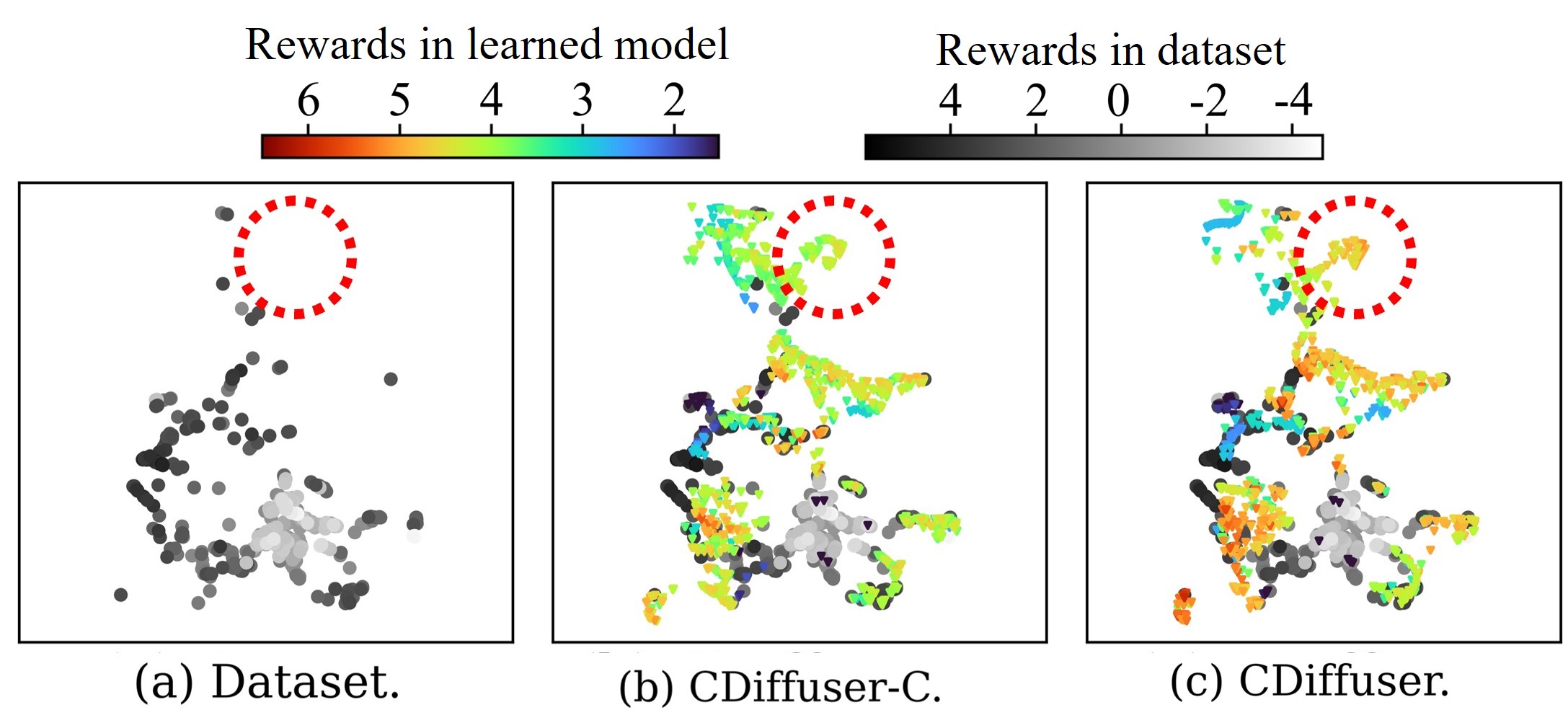}  % 图片路径
    \vspace{-1pt}
    \caption{The distribution of state and reward. It is better to view in color mode. CDiffuser achieves higher rewards in out-of-distribution areas (circled with red).}
    \label{fig: visual1}
  \end{minipage}
  \hfill
  \begin{minipage}[b]{0.43\textwidth}
    \centering
    \includegraphics[width=\textwidth]{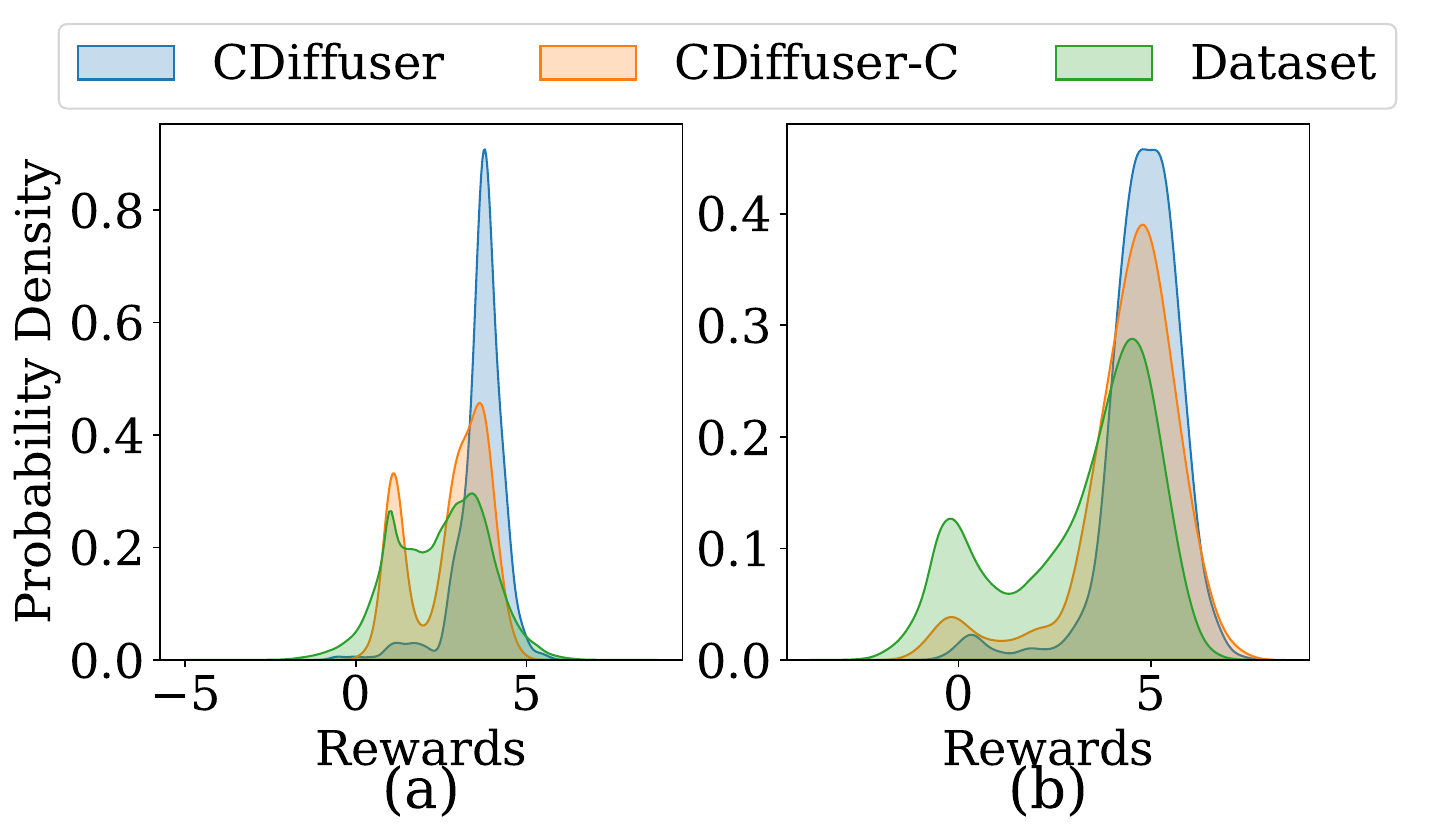}  % 图片路径
    \vspace{-20pt}
    \caption{Distribution of rewards on  (a)  Walker2d-Med-Replay and (b) Halfcheetah-Med-Replay. }
    \label{fig:based}
  \end{minipage}
\end{figure}

% \begin{wrapfigure}{r}{0.5\textwidth}
%     \includegraphics[width=0.48\columnwidth]{figs/revised_visual.py.pdf} 
%     \caption{Distribution of rewards on Walker2d-Med-Replay (a) and Halfcheetah-Med-Replay (b).  CDiffuser has a higher probability density on high rewards in both
%     cases.
%     % The base distribution of Diffuser expresses a higher similarity to the dataset distribution, while CDiffuser concentrates more on the high reward area.
%     }
%     \label{fig:based}
% \end{wrapfigure}

%(1) \textbf{Improving the base distribution benefits the performance.} CDiffuser-G outperforms Diffuser-G in 8 out of 9 tasks. Given that both CDiffuser-G and Diffuser-G eliminate guidance and generate trajectories with their base distributions, this suggests the base distribution of CDiffuser-G leads an improved performance, \ie, improving the base distribution is beneficial.
% eliminate guidance and sample form the base distribution directly,
% the contrastive mechanism implemented in CDiffuser-G enhances the base distribution, leading to improved overall performance.
(1) \textbf{The contrastive mechanism benefits the performance.}  
CDiffuser-C, which eliminates the contrastive mechanism from CDiffuser, exhibits poorer performance across all nine tasks than CDiffuser. This suggests that the contrastive mechanism indeed provides benefits.
% CDiffuser surpasses CDiffuser-C, which removes contrastive from CDiffuser in all nine tasks, intuitively demonstrates the benefits of contrasting the contrastive mechanism. 
%This can be concluded from the comparison between \textit{CDiffuser and CDiffuser-C}, as well as \textit{CDiffuser-G and Diffuser-G}. CDiffuser surpasses CDiffuser-C in all nine tasks, which intuitively demonstrates the clear benefits of contrasting the trajectory generation process with high-value and low-value samples. As CDiffuser-G and Diffuser-G remove guidance from the sampeling process, in other words, their performance stand for the performance of their base distributions. Observing that CDiffuser-G outperforms Diffuser-G in 8 out of 9 tasks, we conclude that CDiffuser improves the base distribution.
% 被注释掉的这一段似乎可以放在further中另起一点。因为我们集中于高return区域，可能导致对低value区域的估计不准确，这可能是个潜在的风险。
% 
% \textbf{CDiffuser concentrates the probability density more on the high-value states without dropping the attention on the low-value states.} Recall that we adopt contrastive learning in CDifffuser to  concentrates the base  distribution of diffusion model to the high-value states. We have demonstrated the improvement of  CDiffuser, yet it is doubtful that this may result in the model being unable to fully learn the distribution of the dataset. To validate the influence of contrastive learning in modeling dataset distribution, we compare \textit{CDiffuser with CDiffuser-N}
(2) \textbf{Applying only the high-return samples in training  diminishes benefits in some cases.} 
Compared with CDiffuser, CDiffuser-N is trained solely with high-return samples. Surprisingly, it achieves a lower performance than CDiffuser in all 9 tasks, indicating that low-return trajectories offer beneficial information for model training.

% relying exclusively on high-return samples brings minimal benefits.

% \begin{wrapfigure}{r}{0.5\textwidth}
%     \includegraphics[width=0.48\columnwidth]{figs/revised_visual.py.pdf} 
%     \caption{Distribution of rewards on Walker2d-Med-Replay (a) and Halfcheetah-Med-Replay (b).  CDiffuser has a higher probability density on high rewards in both
%     cases.
%     % The base distribution of Diffuser expresses a higher similarity to the dataset distribution, while CDiffuser concentrates more on the high reward area.
%     }
%     \label{fig:based}
% \end{wrapfigure}

\paragraph{State-reward distribution analysis.}
To illustrate the advantages of CDiffuser more intuitively, we randomly collect the (state, reward) pairs from the offline dataset of Walker2d-Med-Replay and the (state, reward) pairs collected when CDiffuser and CDiffuser-C interact with the environment. 
% Here, we choose Diffuser and Decision Diffuser as both of them apply diffusion to model RL as a sequence generation problem.
The results are shown in \cref{fig: visual1}, in which each scatter represents a state mapped by UMAP~\citep{mcinnes2018umap}, and its color denotes the reward gained in the corresponding state. From the results illustrated in \cref{fig: visual1}, we can observe that: (1) there are more red and yellow dots in \cref{fig: visual1}(c) than (b). That indicates that the model achieves better rewards with contrasting mechanisms; (2) In out-of-distribution states (circled with red), CDiffuser gains higher rewards than CDiffuser-C, which indicates that contrastive mechanism has potential in tackling out-of-distribution issue.

% in both in-distribution states (circled with blue) and out-of-distribution states (circled with red), CDiffuser gains higher rewards, demonstrating the ability of CDiffuser in generating state-action pairs with higher rewards. 

% Recall that we improve the base distribution of diffusion model with contrastive learning to enhance the performance. 
% To evaluate the improvements of CDiffuser on base distribution, we take walker2d-medium-replay and halfcheetah-medium-replay as representative datasets,  randomly 
% sample trajectories with the base distribution of Diffuser and CDiffuser without guidance, interact with the environment, and record the rewards provided by the environment.
% \paragraph{Base distribution analysis.}
\paragraph{Reward distribution analysis.} The contrast mechanism in CDiffuser plays a relatively significant role in the case of out-of-distribution and limited high-return trajectories. We believe that a deeper reason lies in the role of the contrast mechanism in the generation of the diffusion model. According to \ref{sec:model_learn}, three crucial components of CDiffuser are the trajectory generation of the diffusion model, return prediction, and the contrast mechanism. To validate our assumption, we remove the return prediction of CDiffuser and CDiffuser-C (\textit{i.e.}, remove $\mathcal{J}_{\phi}$ from \cref{eq:sample}), leverage them to generate subsequent trajectory. 
Then we apply the actions in the generated trajectory to interact with the environment Walker2d-Med-Replay and HalfCheetah-Med-Replay, and visualize the distribution of reward during the interaction, as shown in Figure~\ref{fig:based}. It can be observed that the CDiffuser has a higher probability density on high rewards in both cases. That indicates the effect of the contrast mechanism essentially increasing the proportion of high-return trajectories generated by the diffusion model. As demonstrated in \cref{eq:sample}, with the support of both generated high-return trajectories and the return predictor, CDiffuser achieves sound performance.

\subsection{Compatibility Study}
\vspace{-1pt}

\begin{wraptable}{r}{0.5\textwidth}
\vspace{-1pt}
    \centering
    \begin{adjustbox}{width=\linewidth} % 将表格宽度设置为页面宽度的一半
       \begin{tabular}{@{}llcc@{}}
        \toprule
        \textbf{Dataset} & \textbf{Environment} & \multicolumn{1}{l}{\textbf{DD$^+$}}  & \multicolumn{1}{l}{\textbf{Diffuser$^+$}}  \\ \midrule
        Med-Expert       & Halfcheetah          & 0            &  3.1$\pm$0.4                              \\ %\midrule
        Med-Expert       & Hopper               & 5.4$\pm$1.2   & 9.1$\pm$1.2                                             \\ %\midrule
        Med-Expert       & Walker2d             & 6.5$\pm$0.9  &  1.3$\pm$0.4                                            \\ \bottomrule
        \end{tabular}
    \end{adjustbox}
    \caption{The improvements of the normalized score after transplanting the contrastive mechanism of CDiffuser to Decision Diffuser (DD$^+$) and Diffuser (Diffuser$^+$), with $\pm$ denoting the variance.}
    \label{tab:cdd}
\end{wraptable}

% Recall that we design the contrastive mechanism to improve the unguided distribution for better performance. This mechanism can be easily transplanted onto other diffusion-based methods, such as Decision Diffuser~\citep{ajay2023is}.
As we discussed in \cref{sec:base model}, our CDiffuser is build based on Diffuser. To validate the compatibility of the contrastive mechanism of CDiffuser,
we transplant it to Decision Diffuser (DD) ~\citep{ajay2023is}, and evaluate and compare the improvement on three environments.
% with the improvement of contrastive Diffuser, 
The improvements are summarized in \cref{tab:cdd}, in which DD$^+$ and Diffuser$^+$ denote the improvement of introducing contrast mechanism in DD and Diffuser correspondingly.
% We name the contrastive version of Decision Diffuser as CDD. 
As we can observe, DD$^+$ achieves noticeable improvement in 2 out of 3 tasks and Diffuser$^+$ gains improvement in 3 out of 3 tasks, which demonstrates the portability of the contrast mechanism of CDiffuser.
Interestingly, DD$^+$ is unable to achieve any improvement in Halfcheetah-Med-Expert. 
This could be attributed to the separated training (sampling) of states and actions in DD, which results in a failure to effectively model their joint distribution.

\vspace{-10pt}
\section{Related Works}
\vspace{-10pt}
\subsection{Diffusion for Decision-Making}
\vspace{-10pt}
% The strong generative ability of diffusion models opens new perspectives for offline reinforcement learning, and there have been some pioneering works introducing diffusion models into offline reinforcement learning.

% Diffusion models~\citep{ho2020denoising,songscore,songdenoising} are recently proposed generative models, which has good generalization ability and has achieved remarkable results in many fields, such as computer vision~\citep{kim2022diffusionclip,rombach2022high,avrahami2022blended}, natural language processing~\citep{li2022diffusion,saharia2022photorealistic} and recommendation systems~\citep{long2021social}. Some researchers have noticed the generative capabilities of diffusion models, and have introduced them into the field of offline reinforcement learning mainly in three ways.

% One common pipeline of combining diffusion models and RL is adopt them as an sample augmentation tool. SYNTHER~\citep{lu2023synthetic} trains a diffusion with a limited initial dataset, and use which to generates synthetic samples as is real data to update the policy network. Focusing on multi-objective reinforcement learning, AdaptDiffuser~\citep{liang2023adaptdiffuser} generate synthetic trajectorys using the learned diffusion model at each iteration under the guidance of task-related reward function. By replacing the reward function, diffusion model generates samples of various tasks for updating the policy network.

We group the diffusion-based methods in RL into action generation methods and trajectory generation methods. The action generation methods~\citep{ada2023diffusion,wang2022diffusion,chen2022offline,chi2023diffusion} adopt diffusion models as policies to predict the action of the current step. One of the typical works in this group is Diffusion Q-learning~\citep{wang2022diffusion}, which proposes to design the policy as a diffusion model and improve it with double Q-learning architecture. Following Diffusion Q-Learning, SRDPs~\citep{ada2023diffusion} incorporates state reconstruction feature learning into the recent category of diffusion policies to address the out-of-distribution generalization problem. The second group of methods generate the subsequent trajectory including the action to take at the current step by diffusion. For instance, Diffuser~\citep{janner2022planning} models trajectories as sequences of state-action pairs. Based on Diffuser, Decision Diffuser~\citep{ajay2023is} proposes to predict state sequences with a diffusion model conditioned on historical information, and adopts a reverse dynamic model to predict actions based on the generated state sequence. Though these methods have gain significant achievements, they neglect the differences of high-return samples and low-return samples and are limited by their plain base distribution.

\vspace{-8pt}
\subsection{Contrastive Learning in RL}
\vspace{-9pt}
The motivation for introducing contrastive learning in RL is to enrich the representation in the previous works. We group these works into three types. The first type of methods apply contrastive learning to enhance the state representations \citep{laskin2020curl,qiu2022contrastive}. For instance, \citet{laskin2020curl} propose to learn image representations via contrastive learning; \citet{qiu2022contrastive} propose to learn the transition with contrastive learning.
The second type of methods apply contrastive learning to learn the representations of tasks. For instance, \citet{yuan2022robust} apply contrastive learning to enhance the representation of tuples to distinguish between different tasks; \citet{agarwal2020contrastive} apply contrastive learning to learn the representations of the environments.
Some works apply contrastive learning in other ways. For instance, \citet{laskin2022unsupervised} utilize
contrastive learning to learn behavior representations and maximizes the entropy to encourage behavioral diversity.
In contrast to the methods mentioned above, CDiffuser adopts contrastive learning to constrain the generated sample, rather than learning representations.

\vspace{-13pt}
\section{Conclusion and Discussion}
\vspace{-8pt}
\label{conclusion}
In this paper, we introduce CDiffuser for offline RL, a contrastive mechanism that uses low-return trajectories and addresses the challenge of limited high-return trajectories.
% which improves the base distribution by introducing contrastive learning.
Different from the previous works which apply contrastive learning to enhance the representation, we perform contrastive learning over the return of states. Specifically, we apply diffusion to generate the subsequent trajectory for planning, and constrain the states in the generated trajectory toward the states with high returns and away from the states with low returns to improve the base distribution. In that way, the actions taken by the agent are always toward the high-return states, which makes the agent gain better performance in the online evaluation. We evaluate CDiffuser on 14 D4RL benchmarks, where the results demonstrate that our CDiffuser achieves outstanding performance. The ablation studies and investigations further substantiated the rationality of CDiffuser.
% we design cl mechan in ... (states), while can be done on more wide areas such as actiosn.  require further explore.
We currently focus on performing contrastive learning over the return of states to enhance the base distribution in this paper. Yet, this contrastive mechanism can also be applied to other levels, such as action or state-action pairs, or even at the latent of trajectories, which we leave to future work.

% We hope that our work serves as a catalyst, inspiring more researchers to pay attention to and explore the role of the data distribution itself in various ways.

% \shanyx{As we have changed the motivation, Contribution Abs and Conclusion maybe require some modfications, to concentrate on all kinds of diffusion models rather than long-time planning or not.}
% In this paper, we introduce CDiffuser for offline reinforcement learning, which combines contrastive learning with diffusion models and is able to plan states with higher rewards while maintaining long-time planning capabilities. \shanyx{Contribution, Abstract and Conclusion should be the same.}
% Whats more, CDiffuser makes a balance between learning the dataset distribution and targeting high-reward states. 

% Despite than experiments across nine simulation tasks demonstrate that CDiffuser achieves notable results, we further conducted more detailed analyses comparing CDiffuser to another two methods, Diffuser and Decision Diffuser, which adopt similar frameworks. Results show that Contrastive Diffsuer is able to (1) concentrate the generated samples more in the high-reward regions, (2) be more effective in mitigating OOD problem, and (3) provide more accurate long-horizon planning. These experimental results demonstrate that leveraging the implicit information within the dataset through contrastive learning significantly enhances the model's performance across various aspects. 

\bibliographystyle{plainnat}
\bibliography{cdiffuser}

\newpage
% \appendix
\appendix

\section{Appendix}

\subsection{Pseudo code of CDiffuser.}
\label{appendix:psuedo:all}
\label{appendix:psuedo:train}
\begin{algorithm}
\label{psuedocode}
  \caption{Training}
  \begin{algorithmic}[1]
    % \Require input
    % \Ensure output
    \STATE Calculate the candidate set $\mathcal{C}$.
    \WHILE{not converged}
        \STATE $\bm{\tau}_t, v_t \sim \mathcal{D}$.
        \STATE $i \sim [1,N]$.
        \STATE Generate $\bm{\tau}_t^i$.
        %with \todo{---}.
        \STATE Reconstruct $\bm{\tau}_t$ as $\hat{\bm{\tau}}^{i,0}_t$ = $\psi_\theta(\bm{\tau}^i_t,i)$.
        \STATE Calculate loss $\mathcal{L}_d$ with \cref{eq:loss:difusion}.
        % \STATE Predict return-to-go with $\tau^i$ using \cref{eq:pred:y}.
        \STATE Calculate loss $\mathcal{L}_v$ with \cref{eq:loss:y}.
        \STATE Extract STATEs in $\hat{\bm{\tau}}_t^{i,0}$ as $\mathcal{S}_{\hat{\bm{\tau}}^{i,0}_t} = \{\hat{\bm{s}}_{t+1}^{i,0},\hat{\bm{s}}_{t+2}^{i,0},...,\hat{\bm{s}}_{t+H}^{i,0}\}$.
        \FOR{$\hat{s}_h^{i,0}$ in $\mathcal{S}_{\hat{\bm{\tau}}^{i,0}_t}$}
            \STATE Sample $\mathcal{S}^{+}$ and $\mathcal{S}^{-}$ with \cref{sec:posi_nega}.
            \STATE Calculate $\mathcal{L}_{h}^i$ using \cref{eq:nega:hard}.
        \ENDFOR
        \STATE Calculate $\mathcal{L}_{c}$ using \cref{eq:loss:contrast:all}.
        \STATE Calculate $\mathcal{L}$ using \cref{eq:loss:all}.
        \STATE Update model by taking gradient decent with $\mathcal{L}$. 
    \ENDWHILE
    
    % \For{每个元素}
    %   \STATE 执行操作
    % \EndFor
    
    % \STATE \Return 结果
  \end{algorithmic}
\end{algorithm}

\begin{algorithm}
\label{appendix:psuedo:planning}
  \caption{Planning}
  \begin{algorithmic}[1]
    % \Require input
    % \Ensure output
    \REQUIRE CDiffuser $\psi_\theta(\cdot,\cdot)$, return-to-go predictor $\mathcal{J}_{\phi}(\cdot, \cdot)$, guidance scale $\rho$, co-variances $\Sigma^i$.
    \STATE $t \leftarrow 1$. 
    \WHILE{not done}
        \STATE Observe STATE $\bm{s}_t$; sample $\bm{\tau}^N_t \sim \mathcal{N}(\bm{0}, \bm{I})$
        \FOR{$i=N,N-1,...,1$}
            \STATE Predict return-to-go with $\mathcal{J}_\phi(\hat{\bm{\tau}}^i_t, i)$.
            \STATE Sample $\hat{\bm{\tau}}^{i-1}_t$ using \cref{eq:sample}.
        \ENDFOR
        \STATE Extract $\hat{\bm{a}}_t$ form $\hat{\bm{\tau}}^0$.
        \STATE Interact with environment using action $\hat{\bm{a}}_t$.
        \STATE $t \leftarrow t+1$. 
    \ENDWHILE
    
    % \For{每个元素}
    %   \STATE 执行操作
    % \EndFor
    
    % \STATE \Return 结果
  \end{algorithmic}
\end{algorithm}

\begin{figure}
    \centering
    \includegraphics[width=1.0\columnwidth]{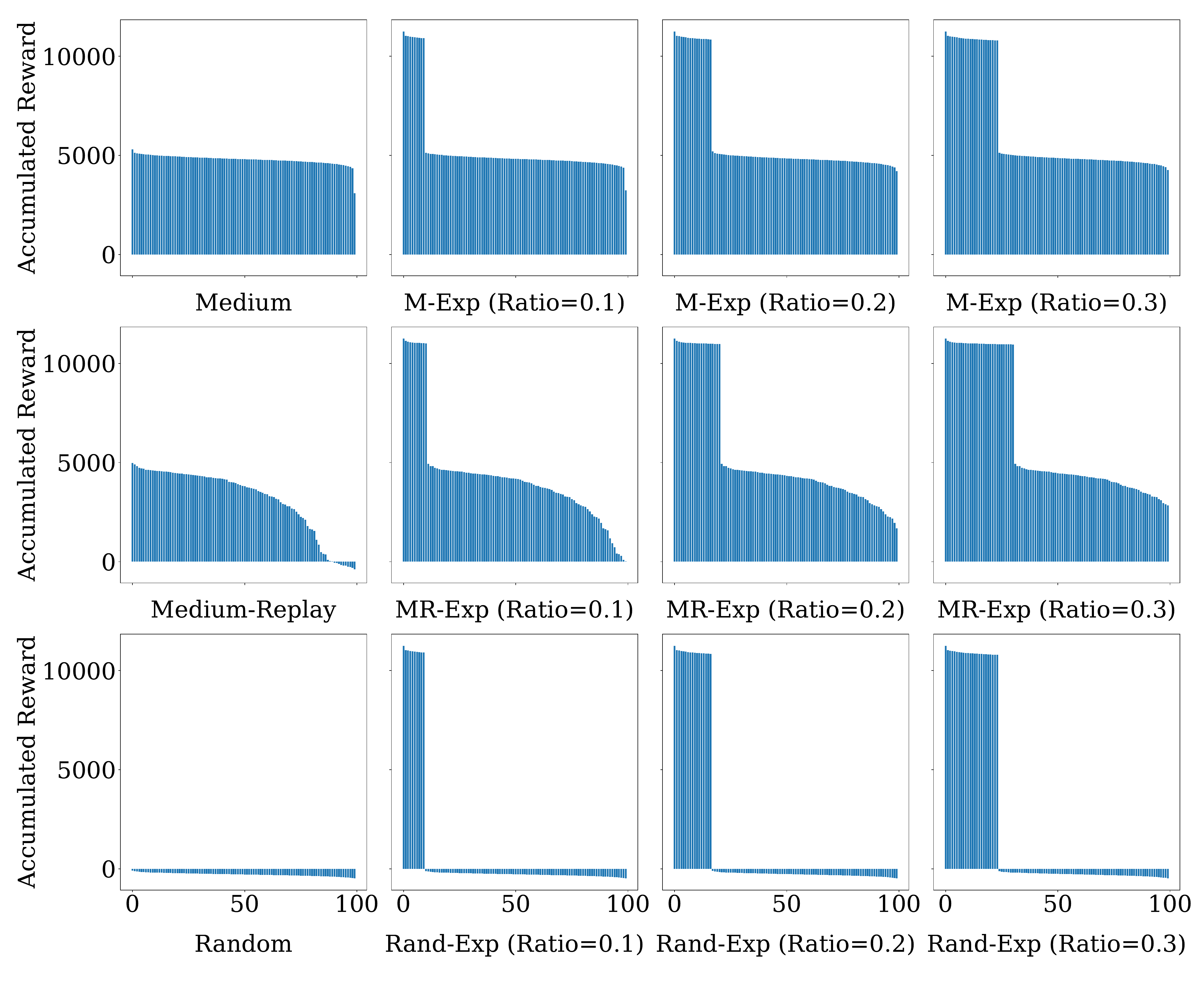}
    \caption{Returns distribution of Medium, Medium-replay and Random datasets of Halfcheetah mixed with different ratios of expert data.}
    \label{fig:mixed_halfcheetah}
\end{figure}

\subsection{Illusion of \cref{eq:posi:hard} and \cref{eq:nega:hard}.}
\begin{figure}
    \centering
    \includegraphics[width=0.9\columnwidth]{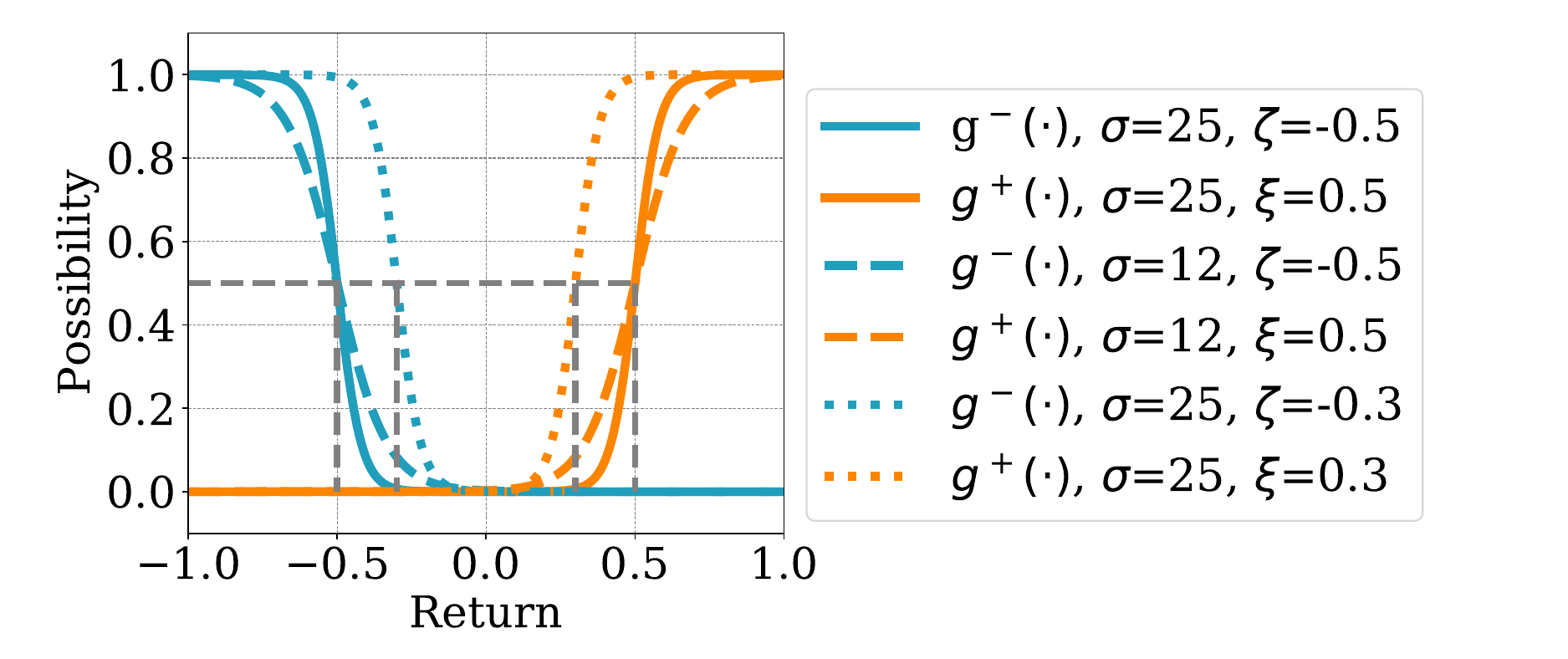}
    \caption{Illusion of \cref{eq:posi:hard} and \cref{eq:nega:hard}}
    \label{fig:g}
\end{figure}

Example of \cref{eq:posi:hard} and \cref{eq:nega:hard} are visualized in \cref{fig:g}. As can be observed in \cref{fig:g}, our modified influence functions are designed to leave a blank in the middle area deliberately, which is different from \citep{thoma2020soft}.  The underlying reason is that not all states are supposed to be contrastive samples. Nevertheless, our modified influence functions collapse into modified influence functions in \citep{thoma2020soft} if we set $\xi=\zeta$.

\subsection{Results of CDiffuser and baseline methods on mixed datasets of various environments.}

We provide the results of CDiffuser and trajectory-based baseline methods ($i.e.$, DT, TT, Diffuser and DD) on mixed datasets of various environments. Our method CDiffuser achieves the optimal and sub-optimal results on 24 out of 27 baseline methods, showing the advantage of CDiffuser on datasets with sparse high-reward samples.

\begin{table}[]
\caption{The average normalized score of different methods on mixed datasets of various environments, with $\pm$ denoting the standard deviation. The mean and standard deviation are computed over 50 random seeds. The best and the second-best results of each setting are marked as \textbf{bold} and \underline{underline}, respectively. \label{tab:all:mix} }
    
\resizebox{1\linewidth}{!}{
\begin{tabular}{@{}clccccc|cc@{}}
\toprule
\multicolumn{1}{l}{\textbf{Environment}} & \multicolumn{1}{l}{\textbf{Dataset}} & \multicolumn{1}{l}{\textbf{Mix Ratio}} & \textbf{DT}             & \textbf{TT}          & \textbf{Diffuser}     & \textbf{DD}              & \textbf{CDiffuser-SR}  & \textbf{CDiffuser-SRD}    \\ \midrule
\multirow{9}{*}{Halfcheetah}             & \multirow{3}{*}{Medium}              & 0.1                                    & 58.6 $\pm$ 2.1          & 46.7 $\pm$ 1.1       & 71.5 $\pm$ 1.8        & 43.2 $\pm$ 0.9           & \underline{ 72.1 $\pm$ 0.9}     & \textbf{73.6 $\pm$ 1.2}   \\
                                         &                                      & 0.2                                    & 46.8 $\pm$ 0.5          & 46.9 $\pm$ 1.8       & \underline{ 80.3 $\pm$ 0.8}  & 41.6  $\pm$ 2.2          & \textbf{81.3 $\pm$ 1.3}  & 77.5 $\pm$ 0.9            \\
                                         &                                      & 0.3                                    & 70.9 $\pm$ 1.2          & 47.4 $\pm$ 1.1       & \underline{ 81.7 $\pm$ 2.9}  & 43.6 $\pm$ 3.3           & \textbf{82.8 $\pm$ 1.2}  & 68.3 $\pm$ 1.1            \\ \cmidrule(l){2-9} 
                                         & \multirow{3}{*}{Med-Replay}          & 0.1                                    & 7.5 $\pm$ 2.1           & \underline{ 42.9 $\pm$ 2.5} & 38.1 $\pm$ 1.1        & 33.8 $\pm$ 1.6           & 37.8 $\pm$ 0.1           & 39.0 $\pm$ 0.5            \\
                                         &                                      & 0.2                                    & 6.7 $\pm$ 4.5           & 43.7 $\pm$ 1.7       & 46.2 $\pm$ 0.2        & 32.8 $\pm$ 0.6           & \underline{ 50.6 $\pm$ 1.2}     & \textbf{58.4 $\pm$ 0.9}   \\
                                         &                                      & 0.3                                    & 6.1 $\pm$ 0.1           & 49.3 $\pm$ 2.2       & \underline{ 57.1 $\pm$ 1.8}  & 36.2 $\pm$ 1.1           & \textbf{60.3 $\pm$ 2.7}  & 55.9 $\pm$ 1.5            \\ \cmidrule(l){2-9} 
                                         & \multirow{3}{*}{Random}              & 0.1                                    & 5.1 $\pm$ 0.0           & 7.7 $\pm$ 0.1        & \underline{ 33.8 $\pm$ 1.8}  & 13.8 $\pm$ 0.8           & 18.1 $\pm$ 1.1           & \textbf{48.0 $\pm$ 2.9}   \\
                                         &                                      & 0.2                                    & 10.3 $\pm$ 4.2          & 16.8 $\pm$ 1.8       & \underline{ 74.4 $\pm$ 1.5}  & 8.5 $\pm$0.2             & 72.3 $\pm$ 0.7           & 65.0 $\pm$ 1.3            \\
                                         &                                      & 0.3                                    & 27.5 $\pm$ 6.8          & 5.9 $\pm$ 0.2        & 75.8 $\pm$ 2.3        & 13.9 $\pm$ 1.1           & \underline{ 86.6 $\pm$ 1.8}     & \textbf{88.7  $\pm$ 0.9}  \\ \midrule
\multirow{9}{*}{Hopper}                  & \multirow{3}{*}{Medium}              & 0.1                                    & 27 $\pm$ 4.3            & 45.2 $\pm$ 1.2       & 82.3 $\pm$ 1.7        & 85.8 $\pm$ 0.7           & \underline{ 87.1 $\pm$ 1.2}     & \textbf{93.3 $\pm$ 1.1}   \\
                                         &                                      & 0.2                                    & 24.9 $\pm$ 2.0          & 45.7 $\pm$ 0.8       & 89.4 $\pm$ 1.5        & 90.1 $\pm$ 0.1           & \underline{ 97.9 $\pm$ 0.9}     & \textbf{100.1 $\pm$ 0.9}  \\
                                         &                                      & 0.3                                    & 20.6 $\pm$ 3.1          & 51.3 $\pm$ 1.3       & 104.8 $\pm$ 0.4       & 96.3 $\pm$ 0.6           & \underline{ 106.0 $\pm$ 2.9}    & \textbf{106.0  $\pm$ 1.4} \\ \cmidrule(l){2-9} 
                                         & \multirow{3}{*}{Med-Replay}          & 0.1                                    & 48.2 $\pm$ 0.9          & 29.7 $\pm$ 0.9       & 63.2 $\pm$ 0.9        & \underline{ 78.8 $\pm$ 1.2}     & \textbf{80.3 $\pm$ 2.6}  & 54.8 $\pm$ 1.9            \\
                                         &                                      & 0.2                                    & 46.6 $\pm$ 1.5          & 31.5 $\pm$ 4.3       & \underline{ 69.5 $\pm$ 3.2}  & 69.4 $\pm$ 4.3           & \textbf{73.9 $\pm$ 0.1}  & 51.7 $\pm$ 0.8            \\
                                         &                                      & 0.3                                    & 55.2 $\pm$ 0.5          & 28.1 $\pm$ 1.9       & \underline{ 69.7 $\pm$ 1.4}  & \textbf{105.4 $\pm$ 3.2} & 65.4 $\pm$ 1.1           & 67.8 $\pm$ 1.9            \\ \cmidrule(l){2-9} 
                                         & \multirow{3}{*}{Random}              & 0.1                                    & \underline{ 51.2 $\pm$ 1.9}    & 2.1 $\pm$ 0.1        & 33.4 $\pm$ 1.4        & 0.9 $\pm$ 1.1            & \textbf{52.0 $\pm$ 0.7}  & 46.9 $\pm$ 3.3            \\
                                         &                                      & 0.2                                    & 48.9 $\pm$ 2.1          & 2 $\pm$ 0.0          & 63 $\pm$ 0.4          & 1.1 $\pm$ 0.4            & \underline{ 67.0 $\pm$ 2.2}     & \textbf{75.3 $\pm$ 1.0}   \\
                                         &                                      & 0.3                                    & 70.9 $\pm$ 1.5          & 2.1 $\pm$ 0.0        & 70.5 $\pm$ 1.2        & 1.8 $\pm$ 0.1            & \textbf{86.4 $\pm$ 1.5}  & \underline{ 83.3 $\pm$ 1.2}      \\ \midrule
\multirow{9}{*}{Walker2d}                & \multirow{3}{*}{Medium}              & 0.1                                    & 91.0  $\pm$ 1.0         & 82.1 $\pm$ 2.1       & \underline{ 93.3 $\pm$ 0.6}  & 49.9 $\pm$ 0.3           & 81.4 $\pm$ 0.4           & \textbf{107.1 $\pm$ 1.6}  \\
                                         &                                      & 0.2                                    & 108.9 $\pm$ 0.2         & 82 $\pm$ 1.7         & \underline{ 102.9 $\pm$ 1.7} & 56.2 $\pm$ 2.9           & \textbf{103.1 $\pm$ 1.2} & 82.0 $\pm$ 0.2            \\
                                         &                                      & 0.3                                    & 37.2 $\pm$ 0.5          & 81.5 $\pm$ 1.2       & 95.21 $\pm$ 0.3       & 31.7 $\pm$ 3.1           & \textbf{102.3 $\pm$ 1.5} & \underline{ 97.0 $\pm$ 2.2}      \\ \cmidrule(l){2-9} 
                                         & \multirow{3}{*}{Med-Replay}          & 0.1                                    & 64.3 $\pm$ 1.9          & 45.2 $\pm$ 1.1       & \underline{ 84.0 $\pm$ 1.0}  & 66.8 $\pm$ 2.2           & \textbf{90.5 $\pm$ 2.2}  & 61.4 $\pm$ 1.2            \\
                                         &                                      & 0.2                                    & 21.8 $\pm$ 4.4          & 41.21 $\pm$ 2.0      & 83.3 $\pm$ 0.7        & \underline{ 84.6 $\pm$ 1.4}     & \textbf{90.8 $\pm$ 0.6}  & 75.9 $\pm$ 0.7            \\
                                         &                                      & 0.3                                    & 37.2 $\pm$ 2.5          & 17.1 $\pm$ 1.2       & \underline{ 86.9 $\pm$ 2.2}  & 70.6 $\pm$ 1.2           & \textbf{93.2 $\pm$ 0.3}  & 80.4 $\pm$ 1.1            \\ \cmidrule(l){2-9} 
                                         & \multirow{3}{*}{Random}              & 0.1                                    & 10.5 $\pm$ 0.1          & 5.1 $\pm$ 0.1        & 14.6 $\pm$ 0.8        & 0                        & \textbf{20.2 $\pm$ 1.3}  & \underline{ 14.7 $\pm$ 1.9}      \\
                                         &                                      & 0.2                                    & \textbf{89.1 $\pm$ 0.9} & 5.6 $\pm$ 0.7        & 48.8 $\pm$ 3.2        & 55.2 $\pm$ 1.9           & \underline{ 57.4 $\pm$ 0.7}     & 51.0 $\pm$ 0.9            \\
                                         &                                      & 0.3                                    & \textbf{85.3 $\pm$ 3.2} & 3.9 $\pm$ 5.4        & 48.9 $\pm$ 2.9        & 60.1 $\pm$ 3.1           & \underline{ 78.4 $\pm$ 1.2}     & 48.3 $\pm$ 2.2            \\ \bottomrule
\end{tabular}
}
\end{table}

\subsection{Which plan to select for the construction of contrastive sample?}
\label{appendix:choice}
We provide two implementations to construct contrastiev samples, namely CDiffuser-SR and CDiffuser-SRD. As described in \cref{sec:posi_nega}, CDiffuser-SR pulls the generated trajectories towards the globally high-return states, while CDiffuser-SRD pulls the generated trajectories towards the transitionable high-return states. This may result in a slight decrease in comparative effectiveness, but it ensures dynamic consistency among contrastive samples.

\begin{figure}[htbp]
    \centering
    \begin{minipage}[t]{0.45\textwidth}
        \centering
        \includegraphics[width=\textwidth]{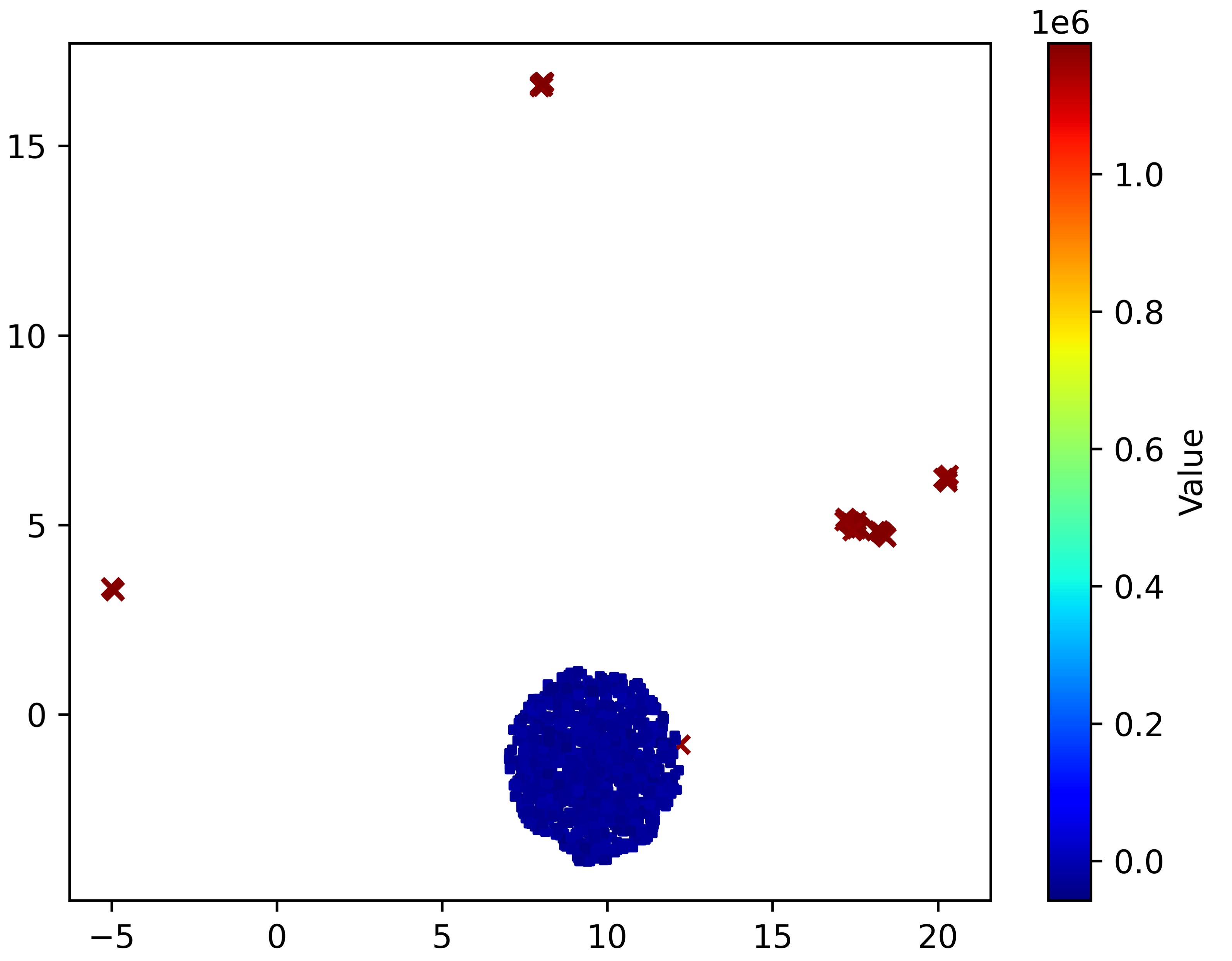} % 第一张图片
        % \caption{(a) Halfcheetah-Random-0.1}
        % \label{fig:select:image1}
    \end{minipage}
    \hfill
    \begin{minipage}[t]{0.5\textwidth}
        \centering
        \includegraphics[width=\textwidth]{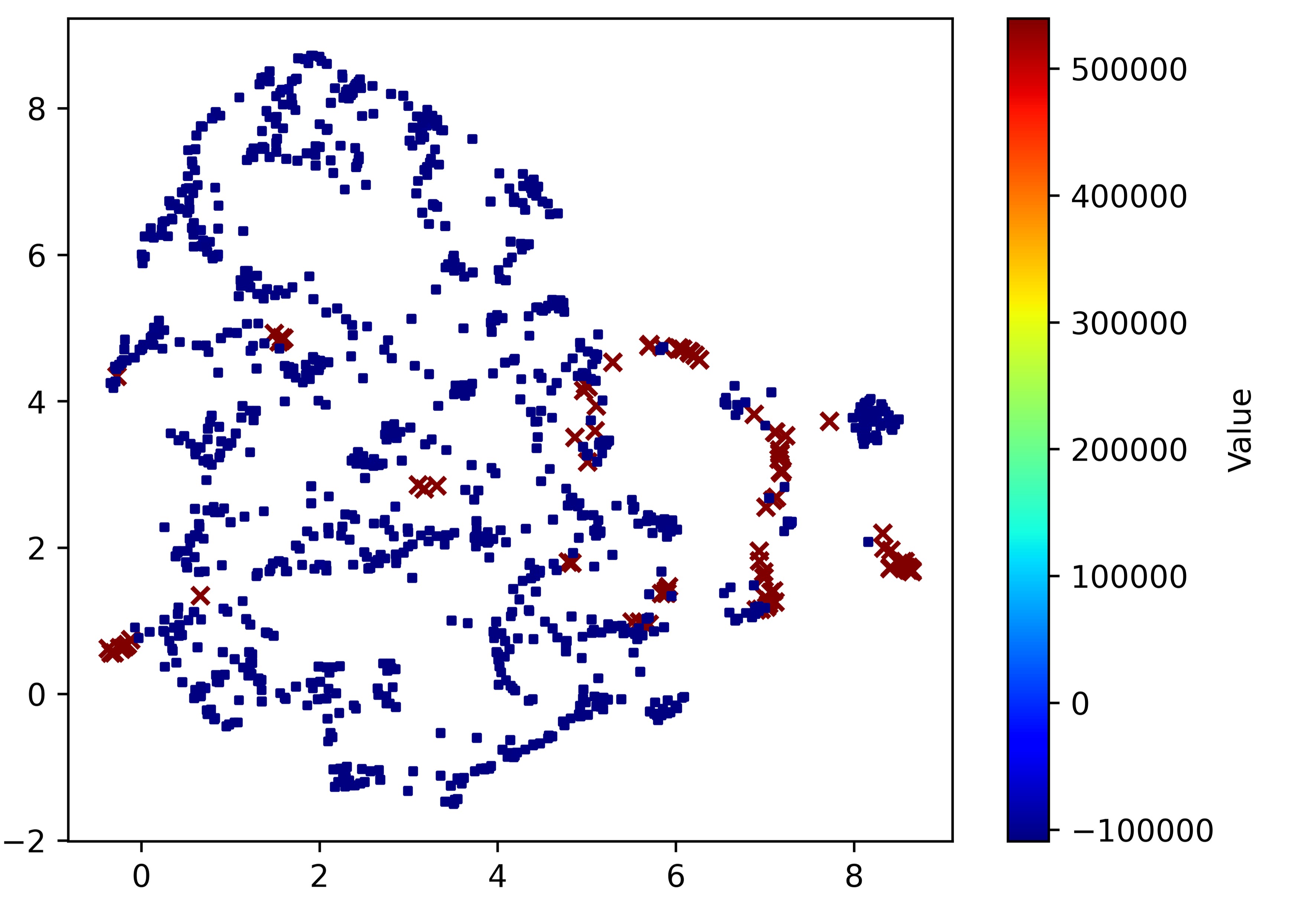} % 第二张图片
        % \caption{(b) Walker2d-Random-0.3}
        % \label{fig:select:image2}
    \end{minipage}
    \caption{States distribution of (a) Halfcheetah-Random-0.1 and (b) Walker2d-Random-0.3, with cross representing the expert data.}
    \label{fig:select:whole}
\end{figure}

Although the results indicate that each method has its advantages, there are also patterns to follow when making a choice. We first visualize the samples and then determine which implementation to choose based on the visualization results. For example, let's consider Halfcheetah-Random-0.1 and Walker2d-Random-0.3. The states with their returns are visualized in \cref{fig:select:whole}, where crosses represent the expert data. From this visualization, we can obtain some prior information about dynamic consistency. 

As shown in \cref{fig:select:whole} (a), most of the high-return states are far away from the low-return states. From this observation, we can infer that starting from any one of the original states from the Halfcheetah-Random dataset, it is difficult to transition to the expert data, i.e., the high-return states marked with a cross. Under this situation, pulling the generated trajectories with CDiffuser-SR will introduce uncertainty, as the corresponding state transitions are unachievable. Therefore, we adopt CDiffuser-SRD for Halfcheetah-Random-0.1. 

In contrast, as shown in \cref{fig:select:whole} (b), the mixed expert data of Walker2d-Random-0.3 shares a similar distribution pattern as the original states from the Walker2d-Random dataset. In this situation, we adopt CDiffuser-SR for more direct constrain.

\subsection{Plan-execution consistency analysis.}

\begin{figure*}[t]
\vspace{-6pt}
\centering
\includegraphics[width=0.9\linewidth]{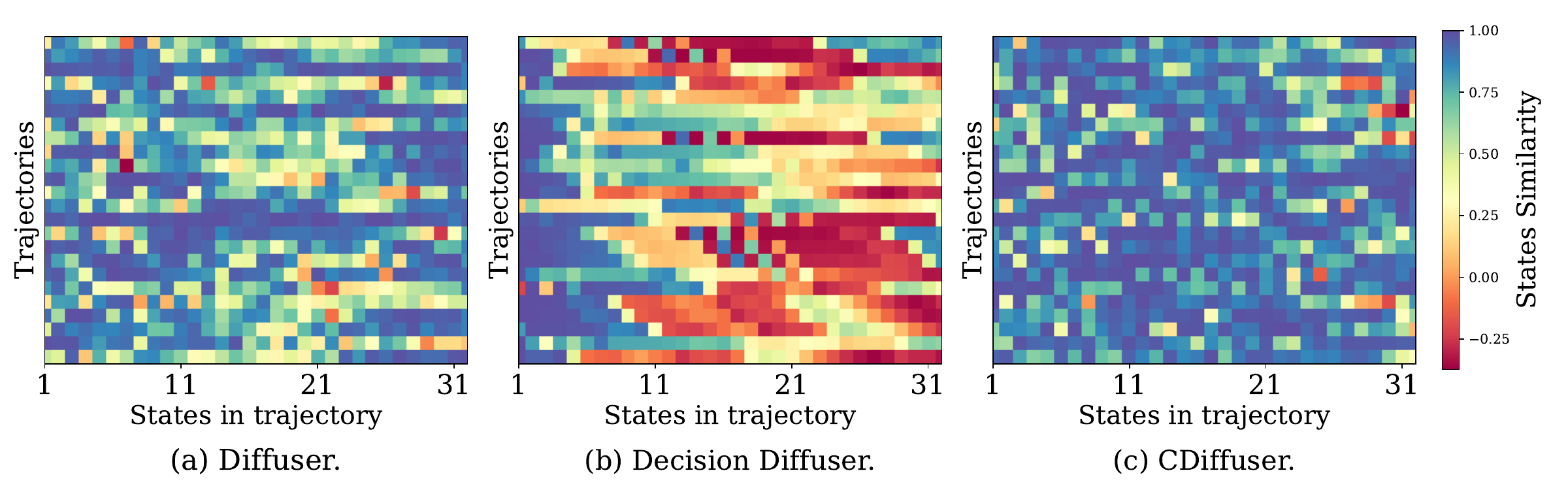} 
\vspace{-10pt}
\caption{The similarities between the states in the generated trajectories and actual states. The generated states of CDiffuser are more similar with the actual states, demonstrating the better long-term dynamic consistency.
}
\vspace{-15pt}
\label{fig: visual2}
\end{figure*}

We use the plan-execution consistency to denote the similarity between the states in the planned trajectory and the states encountered by the agent during its interaction with the environment. It reflects the models' capability in modeling the environment.
% The plan-execution consistency, which denotes the similarity between the states in the planned trajectory and the states encountered by the agent during its interaction with the environment, potentially reflects the models' capability in modeling the environment. 
To investigate the plan-execution consistency of CDiffuser, we randomly take 24 trajectories generated by Diffuser, Decision Diffuser, and CDiffuser. For each generated trajectory, we take the states of consecutive 32 steps and compute the similarity between each generated state and the actual state of the same step provided by the environment. Thus, there are $24 \times 32$ similarity returns for each model, which corresponds to a similarity matrix as the subgraphs in Figure \ref{fig: visual2} illustrated. Each line in the subgraphs of Figure \ref{fig: visual2} represents a generated trajectory, and the grids of each line represent the similarity of the states in the generated trajectory and the states provided by the environment. From Figure \ref{fig: visual2}, we can observe that: (1) Most grids in Figure \ref{fig: visual2} (c) are blue, which denotes that most generated states are consistent with the actual states; (2) Figure \ref{fig: visual2} (c) contains more blue grids than Figure \ref{fig: visual2} (a) and (b), which denotes that CDiffuser has better plan-execution consistency than Diffuser and Decision Diffuser. 
Since the difference between CDiffuser and Diffuser is the contrastive module, combining Figure \ref{fig: visual1} and Figure \ref{fig: visual2}, we can conclude that the contrative module benefits the plan-execution consistency of CDiffuser and makes CDiffuser gain high rewards in both in-distribution and out-of-distribution situations.

\subsection{Visualization of positive and negative samples.}
\begin{figure*}[ht]
\vspace{-20pt}
  \centering
  \subfigure[Agents with high-return states.]{\includegraphics[width=1.0\textwidth]{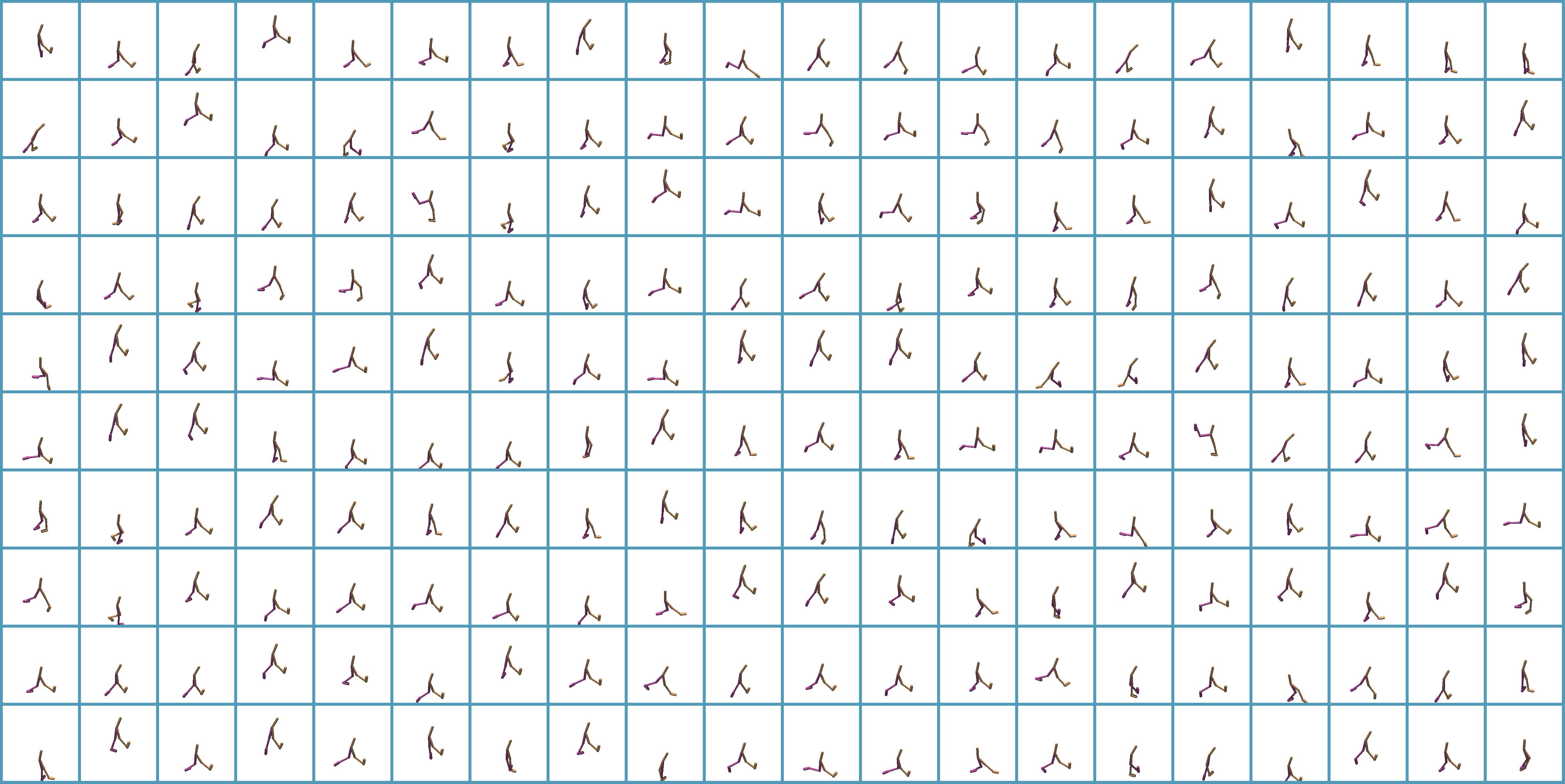}}
  \vspace{1cm}  % 垂直间距
  \subfigure[Agents with low-return states.]{\includegraphics[width=1.0\textwidth]{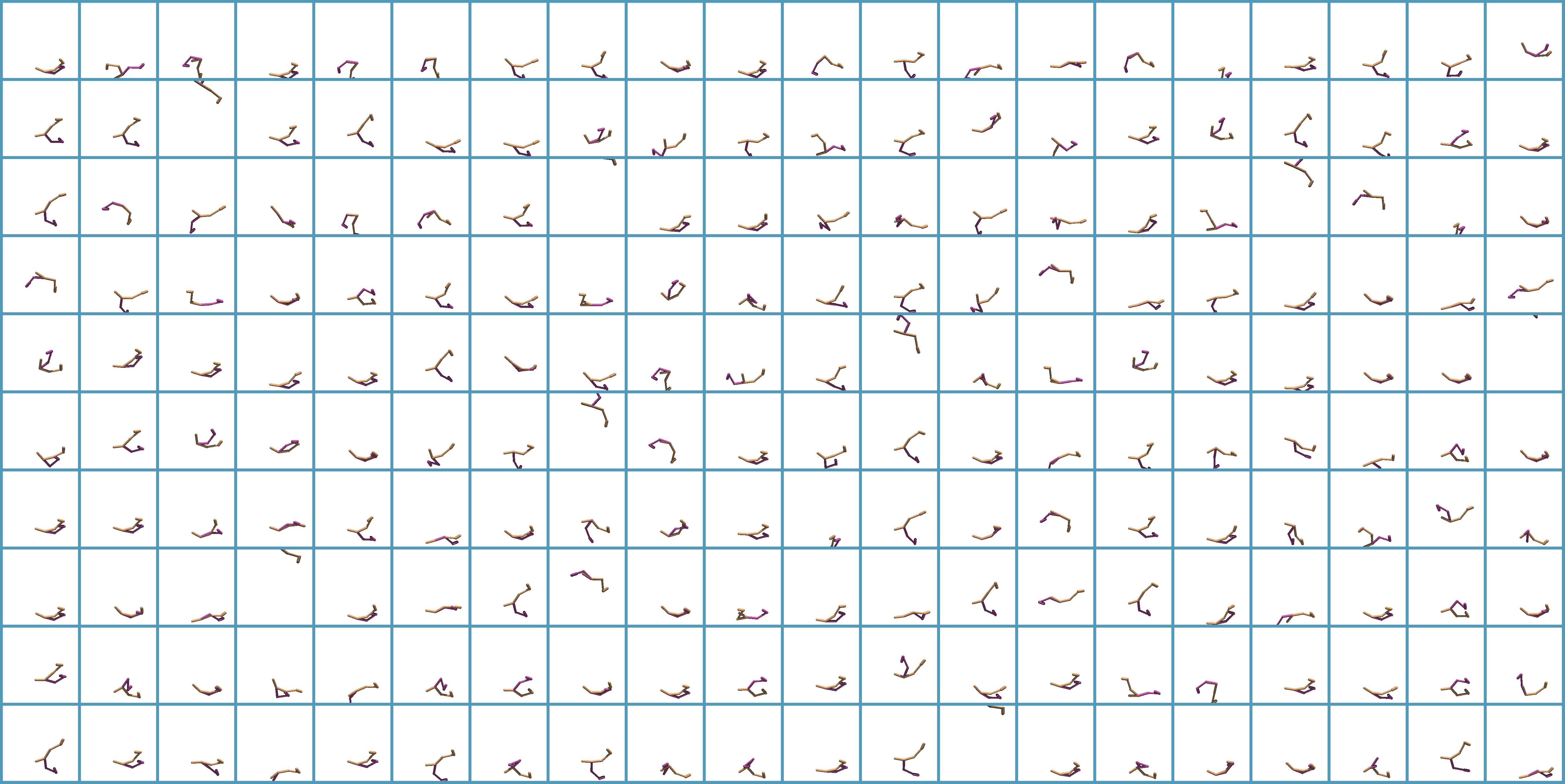}}

  \vspace{-30pt}
  \caption{Visualization of positive samples (states with high returns) and negative samples (states with low returns) in Walker2d-Med-Replay.}
  \label{fig:posi_nega_STATEs}
\end{figure*}

We randomly sample a subset of positive samples (states with high returns) and negative samples (states with low returns), as is shown in \cref{fig:posi_nega_STATEs}. It can be observed that an agent in a state corresponding to a high return tends to be in a position more conducive to walking or running, such as standing upright; correspondingly, an agent with a state corresponding to a low return will be in a position that is hard to walk, such as having already fallen down or about to fall down. This is reasonable, since poses such as standing upright are more conducive to walking or running, which causes the agent to continue moving and results in a higher return, while poses such as having fallen or about to fall cause the environment to give a stop signal, which results in a lower return.

\subsection{Optimizing $\mathcal{J}_\phi(\cdot, \cdot)$ with \cref{eq:loss:all}}
\label{appendix:sec:loss}
Suppose we have the diffuison model $\psi_{\theta}(\cdot)$ parameterized by $\theta$, and the return predictor $\mathcal{J}_{\phi}$ parameterized by $\phi$.
Following \cref{eq:loss:all}, we have

\begin{equation}
    \mathcal{L} = \lambda_d \mathcal{L}_d + \lambda_v \mathcal{L}_v + \lambda_c \mathcal{L}_c.
\end{equation}

Further, 

\begin{equation}
    \mathcal{L}_d = \mathbb{E}_{{\tau}_t\in \mathcal{D}, t>0, i \sim [1, N]}\left[\| {\tau}_t - \psi_\theta({\tau}^i_t,i) \|^2\right],
\end{equation}

\begin{equation}
    \mathcal{L}_v = \mathbb{E}_{{\tau}_t\in \mathcal{D}, t>0, i \sim [1, N]}[\| \mathcal{J}_\phi({\tau}_t^i, i) - v_t\|^2].
\end{equation}

The training process can be viewed as a procedure of calculating gradients of all the parameters and updating them, specifically,

\begin{align}
    \nabla \theta 
    &= \frac{\partial \mathcal{L}}{\partial \theta} \\
    &= \lambda_{d}\frac{\partial \mathcal{L}_d}{\partial \theta} + \lambda_v\frac{\partial \mathcal{L}_v}{\partial \theta} + \lambda_c\frac{\partial \mathcal{L}_c}{\partial \theta} \\
    &= \lambda_v\frac{\partial \mathcal{L}_v}{\partial \theta} + \lambda_c\frac{\partial \mathcal{L}_c}{\partial \theta},
\end{align}

\begin{align}
    \nabla\phi 
    &= \frac{\partial \mathcal{L}}{\partial \phi} \\
    &= \lambda_d\frac{\partial \mathcal{L}_d}{\partial \phi} + \lambda_v\frac{\partial \mathcal{L}_v}{\partial \phi} + \lambda_c\frac{\partial \mathcal{L}_c}{\partial \phi}  \\
    &= \lambda_d\frac{\partial \mathcal{L}_d}{\partial \phi}.
\end{align}

Thus, calculating the gradients of $\theta$ with $\mathcal{L}$ is equal to calculate $\theta$ with $\mathcal{L}_d$ and $\mathcal{L}_c$, calculating the gradients of $\phi$ with $\mathcal{L}$ is equal to calculate $\phi$ with $\mathcal{L}_v$, $i.e.$, optimizing the return predictor $\mathcal{J}\phi(\cdot, \cdot)$ with \cref{eq:loss:all} is equal to optimizing it with \cref{eq:loss:y} only.

\clearpage
\newpage
\end{document}